\documentclass{article}
\usepackage{latexsym,url}
\usepackage{amsmath,amsfonts,amssymb}
\usepackage{graphicx}
\newtheorem{theorem}{Theorem}
\newtheorem{lemma}[theorem]{Lemma}

\newtheorem{definition}{Definition}

\begin{document}
\title{A Large Population Size Can Be Unhelpful in Evolutionary Algorithms}
\author{Tianshi Chen$^{1,2}$, Ke Tang$^{2}$, Guoliang Chen$^{2}$, and Xin Yao$^{2,3}$\\\\
{\small$^1$Institute of Computing Technology, Chinese Academy of Sciences}\\
{\small Beijing 100190, China}\\\\
{\small$^2$Nature Inspired Computation and Applications Laboratory (NICAL)}\\
{\small School of Computer Science and Technology}\\
{\small University of Science and Technology of China}\\
{\small Hefei, Anhui 230027, China}\\\\
{\small $^3$Centre of Excellence for Research in Computational Intelligence and
Applications (CERCIA)}\\
{\small School of Computer Science, University of Birmingham}\\
{\small Edgbaston, Birmingham B15 2TT, UK} }

%\begin{frontmatter}
%\title{A Large Population Size Can Be Unhelpful in Evolutionary Algorithms\tnoteref{work}}
%\tnotetext[work]{This work was partially done when Tianshi Chen was a PhD student at Nature Inspired Computation and Applications Laboratory (NICAL), School of Computer Science and Technology, University of Science and Technology of China, Hefei, Anhui 230027, China.}
%\author[1,2]{Tianshi~Chen}
%\author[2]{Ke~Tang}
%\author[2]{Guoliang~Chen}
%\author[2,3]{Xin~Yao}
%\address[1]{Institute of Computing Technology, Chinese Academy of Sciences, Beijing 100190, China.}
%\address[2]{Nature Inspired Computation and Applications
%Laboratory (NICAL), School of Computer Science and Technology, University of Science and Technology of China, Hefei, Anhui 230027, China.}
%\address[3]{Centre of Excellence for Research in Computational Intelligence and Applications (CERCIA), School of Computer Science, University of Birmingham, Edgbaston, Birmingham B15 2TT,UK.}
%%\thanks[1]{Corresponding author (E-Mail: x.yao@cs.bham.ac.uk). This work is partially supported by a National Natural Science Foundation of China grant (No. 60533020), the Fund for Foreign Scholars in University Research and Teaching Programs (Grant No. B07033), the Fund for International Joint Research Program of Anhui Science and Technology Department (No. 08080703016), and an Engineering and Physical Science Research Council grant in UK (No. EP/D052785/1).}
\maketitle
\begin{abstract}
The utilization of populations is one of the most important features of evolutionary algorithms (EAs). There have been many studies analyzing the impact of different population sizes on the performance of EAs. However, most of such studies are based computational experiments, except for a few cases. The common wisdom so far appears to be that a large population would increase the population diversity and thus help an EA. Indeed, increasing the population size has been a commonly used strategy in tuning an EA when it did not perform as well as expected for a given problem. He and Yao \cite{HeYao02} showed theoretically that for some problem instance classes, a population can help to reduce the runtime of an EA from exponential to polynomial time. This paper analyzes the role of population further in EAs and shows rigorously that large populations may not always be useful. Conditions, under which large populations can be harmful, are discussed in this paper. Although the theoretical analysis was carried out on one multi-modal problem using a specific type of EAs, it has much wider implications. The analysis has revealed certain problem characteristics, which can be either the problem considered here or other problems, that lead to the disadvantages of large population sizes. The analytical approach developed in this paper can also be applied to analyzing EAs on other problems.
\end{abstract}
%\begin{keyword}
%runtime analysis, evolutionary algorithms, computational time complexity, evolutionary computation, combinatorial optimization
%\end{keyword}
%\end{frontmatter}

\section{Introduction}
As a crucial characteristic of Evolutionary Algorithm (EA), the utilization of population enables explorations to different parts of the search space via a number of individuals. Although over the past decades most practical EAs employ populations, the rigorous theoretical investigations on the impact of population on evolutionary algorithms were mainly carried out in the recent eight years. Concerning this issue, He and Yao \cite{HeYao02} took one of the first attempts via the comparisons of the mean first hitting times of both $(N+N)$ and $(1+1)$ EAs on a class of multimodal problems derived from the well-known \textsc{OneMax} problem, and the purpose is to demonstrate the impact of population. Later, a number of theoretical investigations have been dedicated to study the first hitting times of EAs with either multiple parents \cite{witt04,Friedrich08,Oliveto08} or offsprings \cite{jan05,Oliveto08}. It is expected that the EAs under investigations, which were known as $(\mu+1)$ and $(1+\lambda)$ EAs, can establish a bridge from analyzing the $(1+1)$ EA to studying $(\mu+\lambda)$ EAs. In the meantime, it is also reported that the recent investigations on EAs with multiple parents and offsprings (e.g., $(N+N)$ EAs) have eventually brought to the community broader perspectives on understanding the behaviors of population-based EAs. Chen \emph{et al.} analyzed the time complexity of $(N+N)$ EA on some well-known unimodal problems \cite{chen09unimodal} and a wide-gap problem \cite{chen10tcs}. Lehre and Yao studied the impact of mutation-selection balance on the performance of $(N+N)$ EA with linear ranking selection on a multimodal problem \cite{lehre09}, but the influence of employing different population sizes was not investigated. It is still not clear how different population sizes influence the performance of $(N+N)$ EA on multimodal problems. In this paper, we carry out theoretical investigations on this issue.

To study the effect of population size via theoretical investigation, it is clear that we need a suitable measure of performance for EAs. Most previous investigations adopt the well-known first hitting time measure, which is a random variable demonstrating the number of generations required by the EA to find the global optimum for the first time. Here we denote by $\tau$ the first hitting time, and $\mathbb{P}(\tau\le a)$ the accumulated probability of $\tau$. We further denote by $\tau_1$ and $\tau_2$ the first hitting times of two arbitrary EAs, say, EA-I and EA-II, respectively. If the following conditions holds
\begin{itemize}
\item $a_1(n)$ is a polynomial function of the problem size $n$;
\item $a_2(n)$ is a super-polynomial function of the problem size $n$;
\item $\mathbb{P}[\tau_1\le a_1(n)]$ is super-polynomially close to $1$;
\item $\mathbb{P}[\tau_2\le a_2(n)]$ is super-polynomially close to $0$,
\end{itemize}
then one can conclude that EA-I is more efficient than EA-II with a probability that is super-polynomially close to $1$. A recent example, which successfully utilizes the above methodology in comparison of different EAs, is provided in \cite{chen09EDA}. However, when considering the following case, direct comparison of the first hitting times would become infeasible:
\begin{itemize}
\item $a_1(n)$ is a polynomial function of the problem size $n$;
\item $a_2(n)$ is a super-polynomial function of the problem size $n$;
\item $a_3(n)$ is a super-polynomial function of the problem size $n$;
\item The reciprocal of $\mathbb{P}[\tau_1\le a_1(n)]$ is bounded from above by some polynomial function of $n$;
\item The reciprocal of $\mathbb{P}[\tau_1  > a_2(n)]$ is bounded from above by some polynomial function of $n$;
\item $\mathbb{P}[\tau_2\le a_3(n)]$ is super-polynomially close to $0$.
\end{itemize}
The reason is that EA-I is likely to perform as inefficient as EA-II. Nevertheless, since EA-I still takes a relatively high probability to perform efficient while EA-II performs inefficiently almost surely, we can still compare the performances of EA-I and EA-II by employing the \emph{solvable rate} as an alternative measure, where the solvable rate is the probability that the EA finds the global optimum of an optimization problem within a polynomial number of generations. The solvable rate can be considered as a generalized measure based on the probability distribution of the traditional first hitting time measure, and it concerns more about the probability with which an EA performs efficiently in general, rather than the detailed computation time for finding the optimum. In fact, in previous investigations (e.g., \cite{Oliveto07}), the idea of solvable rate has been utilized implicitly in company with the first hitting time results, though it has not been adopted as a measure of performance.

By employing the solvable rate measure in this paper, we carry out theoretical analysis to study the impact of the population size on the performance of an $(N+N)$ EA on a multimodal problem. This multimodal problem, which is called the \textsc{TrapZeros} problem, contains a global optimum $(1,\dots,1)$ and a local optimum $(0,\dots,0)$. The attraction basin of the global optimum only consists of solutions with the leading substring made up of $\ln^2n+2$ consecutive 1-bits. To find the global optimum, the EA has to enter its basin of attraction first by resisting the selection pressure which tends to preserve the solutions with leading 0-bits. For the $(N+N)$ EA on the above problem, we consider three cases for the population size $N$, $N=1$, $N=O(\ln n)$ and $N=\Omega(n/\ln n)$, where the well-known $(1+1)$ EA is considered as a special case of $(N+N)$ EA ($N=1$). It is discovered that when the population size is relatively small ($N=1$ or $N=O(\ln n)$), the solvable rate of $(N+N)$ EA is still larger than the reciprocal of some polynomial function of the problem size, which implies that the EA, if running on an appropriate polynomial number of processors simultaneously and independently, can find the global optimum with a polynomial number of generations. However, given a large enough population size $N=\Omega(n/\ln n)$, the solvable rate of the $(N+N)$ EA has dropped to a level that is super-polynomially close to $0$, implying that the EA cannot find the global optimum within a polynomial number of generations, unless one can offer a super-polynomial number of processors for the EA to run on.

The rest of the paper is organized as follows: Section \ref{sec:pre} introduces the algorithm and problem investigated in this paper. Section \ref{sec:approach} presents the mathematical tools utilized in this paper. Section \ref{sec:small} analyzes the $(N+N)$ EA with the population size $N=1$. Section \ref{sec:med} shows the analytical result of the $(N+N)$ EA with the population size $N=O(\ln n)$. Section \ref{sec:large} concerns the $(N+N)$ EA with the population size $N=\Omega(n/\ln n)$. Section \ref{sec:discussion} carries out discussions on the results presented in the previous sections. Section \ref{sec:Conclusion} concludes the whole paper.

\section{Problem and Algorithm}
\label{sec:pre} In this section, we introduce the concrete optimization problem and EA investigated in this paper.
\subsection{Problem}\label{sec:problem}
\begin{figure} [htbg]
\centering
\includegraphics[width=0.7\textwidth]{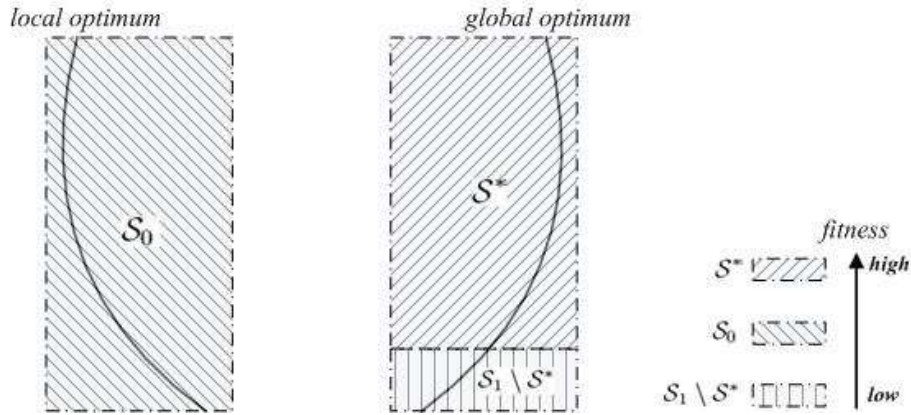}
\caption{Illustration of the \textsc{TrapZeros} problem.} \label{fig:problem}
\end{figure}

The \emph{maximization} problem we consider in this paper, defined over the domain $\textbf{x}=(x_1,\dots,x_n)\in\{0,1\}^n$, is called \textsc{TrapZeros}:
\begin{eqnarray}
\textsc{TrapZeros}(\textbf{x})\triangleq\left\{
\begin{array}{llll}
2n+\sum_{i=1}^{n}\prod_{j=1}^{i}(1-x_{j}), & \text{if } (x_1=0)\wedge(x_2=0); \\
3n+\sum_{i=1}^{n}\prod_{j=1}^{i}x_{j}, & \text{if } (x_1=1)\wedge(x_2=1)\wedge\left(\prod_{i=3}^{\ln^2
n+2}x_i=1\right);\\
n+\sum_{i=1}^{n}\prod_{j=1}^{i}x_{j}, & \text{if } (x_1=1)\wedge(x_2=1)\wedge\left(\prod_{i=3}^{\ln^2
n+2}x_i=0\right);\\
0, & \text{if }(x_1=0)\wedge(x_2=1);\\
1, & \text{if }(x_1=1)\wedge(x_2=0);\\
0, & \text{Other}.
\end{array}
  \right.
\end{eqnarray}
The \textsc{TrapZeros} problem is a multimodal problem, and its global optimum is $\textbf{x}^*=(1,\dots,1)$. For \textsc{TrapZeros}, increasing the leading 0-bits in its solution may eventually lead to the local optimum $(0,\dots,0)$ instead of leading to the global optimum $(1,\dots,1)$. To reach the attraction basin of the global optimum, an optimization algorithm should first find some solutions with the leading substring consisting of $\ln^2n+2$ consecutive 1-bits. Otherwise, the selection pressure of an EA will tend to preserve the solutions with leading 0-bits. To facilitate the later investigations, we define three schemata as follows:
\begin{eqnarray*}
&&\mathcal{S}_1= \{(1,1,*,\dots,*)\},\\
&&\mathcal{S}_0= \{(0,0,*,\dots,*)\},\\
&&\mathcal{S}^*= \left\{(1,1,\underbrace{1,\dots,1}_{\ln^2n},*,\dots,*)\right\},\\
&&\mathcal{S}^*\in\mathcal{S}_1,
\end{eqnarray*}
where ``$*$'' can represent either $0$ or $1$, $\mathcal{S}^*$ and $\mathcal{S}_1$ are the schemata containing the global optimum. Fig. \ref{fig:problem} illustrates the fitness landscape of \textsc{TrapZeros} with respect to the schemata defined above, and it shows that the individuals belonging to $\mathcal{S}^*$ are strictly better than any individual belonging to $\mathcal{S}_0$, while the individuals belonging to $\mathcal{S}_0$ are strictly better than any individual belonging to $\mathcal{S}_1\setminus\mathcal{S}^*$. Utilizing this property, we will carry out rigorous analysis of an EA on \textsc{TrapZeros} later.

\subsection{Algorithm}
The $(N+N)$ EA studied in this paper is with equal parent and offspring sizes. The detailed algorithm is described as follows:
\begin{enumerate}
\item \textbf{Initialization:} The $N$ initial individuals are generated uniformly at random, and the initial
population $\xi_1$ is obtained.
\item \textbf{Mutation:} At the $t^{th}$ generation ($t\in\mathbb{N}^+$), the $N$ individuals in the parent population
$\xi_{t}$ are mutated, and the offspring population $\xi_{t}^{(m)}$ is obtained. The mutation of each individual in $\xi_{t}$ utilizes the {\em bitwise mutation}, i.e., each bit of the individual is flipped independently with a uniform probability $1/n$, where $n$ is the problem size.

\item \textbf{Selection:} After the mutation step at the $t^{th}$ generation ($t\in\mathbb{N}^+$), the best $N$
individuals in the parent and offspring populations ($\xi_{t}\cup\xi_{t}^{(m)}$) are selected to form the population $\xi_{t+1}$, which is the parent population of the $(t+1)^{th}$ generation. Afterwards, set $t=t+1$ and then go to the mutation step.
\end{enumerate}
The execution of the EA will stop if the stopping criterion is met. The above algorithm adopts the truncation selection, and does not employ any recombination operator. The investigation of EAs with recombination operator and other selection operators will be left as our future work.

With respect to the $(N+N)$ EA, the population size $N$ must be a polynomial function of the problem size $n$, otherwise each generation of the EA would require super-polynomial number of fitness evaluations. When $N=1$, the above $(N+N)$ EA degenerates to the well-known $(1+1)$ EA \cite{Droste02}.

\subsection{Solvable Rate}
So far we have introduced the problem and algorithm investigated in this paper. In this subsection, we present the measure of performance for EA. First, let us review the first hitting time $\tau$, which is defined formally as below:
\begin{eqnarray*}
\tau=\min \{t ;x^* \in \xi_{t}\},
\end{eqnarray*}
where $x^*$ is the global optimum, and $\xi_t$ is the population at the $t^{th}$ generation. As the example we have presented in Section I, the first hitting time measure may sometimes become invalid for comparing the performance of two EAs. Here we provide an alternative measure, the solvable rate, to deal with the situation shown in the aforementioned example. Denoted by $\kappa$, the solvable rate is formally defined by
\begin{eqnarray*}
\kappa=\mathbb{P}\left(\tau\prec Poly(n)\right),
\end{eqnarray*}
where the event $\tau\prec Poly(n)$ means that there exists some polynomial function (of the problem size $n$) $F(n)$ such that $\tau< F(n)$ holds for any $n>n_0$, and $n_0$ is a positive constant. Generally speaking, to derive appropriate bounds for the solvable rate, we have to concern the first hitting time $\tau$. In the next section, we introduce the mathematical tools for our further investigations.

\section{Analytical Approaches}\label{sec:approach}
In this section, we present the analytical tools utilized in this paper.
\subsection{Probability Inequalities}
First of all, three well-known probabilistic inequalities are necessary to our later analysis. The inequalities are presented as the following lemmas:
\begin{lemma}[Chernoff bounds \cite{Motwani95book,Droste02}] \label{lemma:chernoff}Let
$X_1,X_2,\dots,X_k \in \{0,1\}$ be $k$ independent random variables with a same distribution:
\begin{eqnarray*}
 \forall i\neq j: \mathbb{P}[X_i=1]=\mathbb{P}[X_j=1],
\end{eqnarray*}
 where $i,j\in\{1,\dots,k\}$. Let $X$ be the sum of those random variables, i.e., $X=\sum_{i=1}^k
 X_i$, then we have
\begin{itemize}
 \item $\forall
 0<\psi<1$: $\mathbb{P}\Big[X<(1-\psi)\mathbb{E}[X]\Big]<e^{-\mathbb{E}[X]\psi^2/2}$.
 \item $\forall 0<\psi\le 2e-1$:
$\mathbb{P}\Big[X>(1+\psi)\mathbb{E}[X]\Big]<e^{-\mathbb{E}[X]\psi^2/4}$.
 \item $\forall \psi>0$:
 $\mathbb{P}\Big[X>(1+\psi)\mathbb{E}[X]\Big]<\Big[\frac{e^\psi}{(1+\psi)^{1+\psi}}\Big]^{\mathbb{E}[X]}$.
\end{itemize}
\end{lemma}

\begin{lemma}[Chebyshev Inequality \cite{Papoulis91book}]
Let $X$ be a random variable with expectation $\mathbb{E}[X]$ and finite variance $Var[X]$. Then for any real number $r
> 0$,
\begin{eqnarray}
\mathbb{P}\bigg[\big|X-\mathbb{E}[X]\big|\ge r\cdot \sqrt{\text{Var}[X]}\bigg]\le \frac{1}{r^2}.
\end{eqnarray}
\end{lemma}

\begin{lemma}[Markov Inequality \cite{Motwani95book}]\label{lemma:markov_inequality}
Let $X\ge0$ be a random variable with expectation $\mathbb{E}[X]$. Then for $a>0$, we have
\begin{eqnarray}
\mathbb{P}[X\ge a]\le \frac{\mathbb{E}[X]}{a}.
\end{eqnarray}
\end{lemma}

\subsection{Decomposition of Population Set}
In this subsection, we introduce the concrete definitions and approaches for analyzing the EA on \textsc{TrapZeros}, which are inherited from our previous investigation \cite{chen09unimodal}. Recall that we have defined the schemata $\mathcal{S}_1$, $\mathcal{S}_0$ and $\mathcal{S}^*$ in Section \ref{sec:problem}. Further, we denote by $E$ the whole population set containing all populations. Based on the aforementioned definitions of the three schemata, we now present the decomposition of the population set $E$, which is necessary for our analytical approaches:
\begin{itemize}
\item We denote by $E_0$ the population set consisting of population $\xi$ with its best individual belonging to
neither $\mathcal{S}_1$ nor $\mathcal{S}_0$.

\item For any population $\xi$ with its best individual belonging to $\mathcal{S}_1$, we define the metric
$m^{(A)}(\xi)$

\begin{eqnarray*}m^{(A)}(\xi)=\min\left\{g^{(A)}(y); \textsc{TrapZeros}(y)=\max\left\{\textsc{TrapZeros}(z); z\in \xi\right\}, y\in\mathcal{S}_1,y\in
\xi\right\},  \end{eqnarray*} where $g^{(A)}(y)=n-2-\sum_{i=3}^{n}|y_{i}-1|$ and $y=(y_1,y_2,\dots,y_n)$ and $z$ are individuals belonging to $\xi$. Based on the metric $m^{(A)}(\xi)$, we obtain $n-1$ population sets, $E^{(A)}_{\rho}$ ($\rho=0,...,n-2$), where
\begin{eqnarray*}
E_\rho^{(A)} =\{\xi;m^{(A)}(\xi) =\rho \}, \quad \rho=0,\dots,n-2.
\end{eqnarray*}
The above definition, along with the fitness function \textsc{TrapZeros}, implies that $E_{n-2}^{(A)}$ is the population set that is made up of all the population containing the global optimum $x^*=(1,1,\dots,1)$.

Moreover, for any population $\xi\in E_{\rho}^{(A)}$, we define a subset of $\xi$:

\begin{eqnarray*}G^{(A)}=\left\{y; \textsc{TrapZeros}(y)=\max\{\textsc{TrapZeros}(z); z\in \xi\}, y\in\mathcal{S}_1,y\in \xi \right\},\end{eqnarray*}

\item For any population $\xi$ with its best individual belonging to $\mathcal{S}_0$, we define the metric
$m^{(B)}(\xi)$

\begin{eqnarray*}m^{(B)}(\xi)=\min\left\{g^{(B)}(y); \textsc{TrapZeros}(y)=\max\left\{\textsc{TrapZeros}(z); z\in \xi\right\}, y\in\mathcal{S}_0,y\in
\xi\right\},  \end{eqnarray*} where $g^{(B)}(y)=n-2-\sum_{i=3}^{n}y_{i}$ and $y=(y_1,y_2,\dots,y_n)$. Based on the metric $m^{(B)}(\xi)$, we obtain $n-1$ population sets, $E^{(B)}_{\rho}$ ($\rho=0,...,n-2$), where \begin{eqnarray*} E_\rho^{(B)} =\{\xi;m^{(B)}(\xi) =\rho \}, \quad \rho=0,\dots,n-2.
  \end{eqnarray*}
Moreover, for any population $\xi\in E_{\rho}^{(B)}$, we define a subset of $\xi$:
\begin{eqnarray*}G^{(B)}=\left\{y; \textsc{TrapZeros}(y)=\max\{\textsc{TrapZeros}(z); z\in \xi\}, y\in\mathcal{S}_0,y\in \xi \right\},\end{eqnarray*}

\item According to the above definitions, we have
\begin{eqnarray*}E=E_0\cup\left(\bigcup_{\rho=0}^{n-2}E_{\rho}^{(A)}\right)\cup\left(\bigcup_{\rho=0}^{n-2}E_{\rho}^{(B)}\right).\end{eqnarray*}
\end{itemize}
On the basis of the decomposition of $E$, we introduce the definitions of local optimal individual (type A and type B):
\begin{definition}[Local Optimal Individual, type A (LOIA)] Given a population set $E^{(A)}_\rho$ ($\rho=0,...,n-2$)
and a population $\xi\in E^{(A)}_\rho$, we call an individual $x\in G^{(A)}$ the $\rho$-LOIA of $\xi$ on $E^{(A)}_\rho$ (LOIA for short, if we restrict the discussion on a given $E^{(A)}_\rho$ and $\xi$), if and only if $x\in G^{(A)}$; We call an individual $x'$ the advanced LOIA for $\forall \xi\in E^{(A)}_\rho$ (advanced LOIA for short, if we restrict the discussion on a given $E^{(A)}_\rho$), if and only if $x'$ is the LOIA of a population $\zeta$ satisfying that $\zeta\in\bigcup_{i=\rho+1}^{n-2}E^{(A)}_i$.
\end{definition}
\begin{definition}[Local Optimal Individual, type B (LOIB)] Given a population set $E^{(B)}_\rho$ ($\rho=0,...,n-2$)
and a population $\xi\in E^{(B)}_\rho$, we call an individual $x\in G^{(B)}$ the $\rho$-LOIB of $\xi$ on $E^{(B)}_\rho$ (LOIB for short, if we restrict the discussion on a given $E^{(B)}_\rho$ and $\xi$), if and only if $x\in G^{(B)}$. We call an individual $x'$ the advanced LOIB for $\forall \xi\in E^{(B)}_\rho$ (advanced LOIB for short, if we restrict the discussion on a given $E^{(B)}_\rho$), if and only if $x'$ is the LOIB of a population $\zeta$ satisfying that $\zeta\in\bigcup_{i=\rho+1}^{n-2}E^{(B)}_i$.
\end{definition}

So far we have defined LOIA and LOIB so as to characterize the individuals belonging to the attraction basins of the global and local optima respectively, which enables us to define the so-called takeover times for the $(N+N)$ EA. Briefly, the takeover time, proposed by Goldberg and Deb \cite{Goldberg91}, originally measures the number of generations required by the population to accumulate enough promising individuals, where a repeated selection process is concerned. Chen \emph{et al.} \cite{chen09unimodal} generalized the concept of ``takeover process'' to characterize the behavior population-based EAs on unimodal problems containing no local optimum. In this paper, on the \textsc{TrapZeros} problem, we further define two types of takeover processes for the global and local optima respectively:

\begin{definition}[$(A,\rho,\epsilon)$-takeover]
A population $\xi$ is said to be $(A,\rho,\epsilon)$-takeover ($\rho=0,\dots,n-2$, and $0<\epsilon\le 1$) if and only if its number of the $\rho$-LOIAs has reached $\lceil\epsilon N\rceil$ (i.e., the proportion of LOIAs in $\xi$ has reached $\epsilon$), where all advanced LOIAs are pessimistically considered as $\rho$-LOIAs.
\end{definition}
\begin{definition}[$(B,\rho,\epsilon)$-takeover]
A population $\xi$ is said to be $(B,\rho,\epsilon)$-takeover ($\rho=0,\dots,n-2$, and $0<\epsilon\le 1$) if and only if its number of the $\rho$-LOIBs has reached $\lceil\epsilon N\rceil$ (i.e., the proportion of LOIBs in $\xi$ has reached $\epsilon$), where all advanced LOIBs are pessimistically considered as $\rho$-LOIBs.
\end{definition}

As it is assumed pessimistically in \cite{chen09unimodal}, throughout the paper we consider that the advanced LOIAs (LOIBs) cannot be generated, if a population is not $(A, \rho,\epsilon)$-taken over ($(B,\rho,\epsilon)$-taken over).Once a population has been $(A, \rho,\epsilon)$-taken over ($(B,\rho,\epsilon)$-taken over), it will concentrate on producing advanced LOIAs (LOIBs) for $E^{(A)}_\rho$ ($E^{(B)}_\rho$). Next, we formally define the $(A,\rho,\epsilon)$-takeover time and $(B,\rho,\epsilon)$-takeover time, and then characterize the processes of generating advanced LOIAs and LOIBs by the so-called $(A,\rho,\epsilon)$ and $(B,\rho,\epsilon)$ upgrade times, respectively.

Given the population $\xi_{t_\rho}\in E^{(A)}_{\rho}$ at the $t_\rho^{th}$ generation, we define its $(A,\rho,\epsilon)$-first hitting time to $E^{(A)}_\rho$ be
\begin{eqnarray}\label{taov}
\eta^{(A)}_{\rho,\epsilon}(\xi_{t_\rho})  = \min\left\{t'_\rho-t_\rho; \xi_{t'_\rho} \in \left(\bigcup_{i=\lceil\epsilon N\rceil}^{N} E^{(A)}_{\rho,i}\right)\cup \left(\bigcup_{j=\rho+1}^{n-2}E^{(A)}_{j}\right)
 \right \},
\end{eqnarray}
where $E^{(A)}_{\rho,i}$ is the population set containing the populations containing $i$ $\rho$-LOIAs ($i=1,\dots,N$). The expectation of $\eta^{(A)}_{\rho,\epsilon}(\xi_{t_\rho})$, restricted to a finite $\eta^{(A)}_{\rho,\epsilon}(\xi_{t_\rho})$ and conditional on the starting population $\xi$, is defined by
\begin{eqnarray*}
 \bar{\eta}^{(A)}_{\rho,\epsilon}(\xi)= \mathbb{E}\left[\eta^{(A)}_{\rho,\epsilon}(\xi_{t_\rho}),\eta^{(A)}_{\rho,\epsilon}(\xi_{t_\rho})<\infty\mid
 \xi_{t_\rho}=\xi\right].
\end{eqnarray*}
The expectation $\bar{\eta}^{(A)}_{\rho,\epsilon}(\xi)$ is called the \emph{\textbf{$(A,\rho,\epsilon)$-takeover time}} of population $\xi$. Afterwards, we define the maximal $(A,\rho,\epsilon)$-takeover time as:
\begin{eqnarray*}\bar{\eta}^{(A)}_{\max,\rho,\epsilon} = \max \left\{\bar{\eta}^{(A)}_{\rho,\epsilon}(\xi); \xi
\in E^{(A)}_{\rho}\right\},
\end{eqnarray*}
and the $(A,\cdot,\epsilon)$-takeover time is defined by
\begin{eqnarray}\label{eq:max_takeover}
\bar{\eta}^{(A)}_{\epsilon}=\max\left\{\bar{\eta}^{(A)}_{\max,\rho,\epsilon}; 0\le \rho\le n-2 \right \}.
\end{eqnarray}
Similarly, we can define $(B,\rho,\epsilon)$-takeover time $\bar{\eta}^{(B)}_{\rho,\epsilon}(\xi)$, and the $(B,\cdot,\epsilon)$-takeover time $\bar{\eta}^{(B)}_{\epsilon}$.

According to the notion presented in \cite{chen09unimodal}, the evolution of individuals, if restricted in a specific attraction basin, can be characterized by the repeated ``takeover-upgrade processes''. Concretely, to reach the optimum in the attraction basin, the $(N+N)$ EA, with both mutation and selection, may often take a number of steps. At each step, the EA may need to accumulate enough promising individuals first, while the qualities of the individuals may not be significantly improved. Afterwards, when a considerable amount of promising individuals have been accumulated, the population will take a high probability for generating one or more better individuals. Formally, we further define the so-called $(A,\rho,\epsilon)$- and $(B,\rho,\epsilon)$- upgrade times. Given the population $\xi_{t_\rho}\in \left(\bigcup_{i=\epsilon N}^{N} E^{(A)}_{\rho,i}\right)$ at the $t_\rho^{th}$ generation, we define its $(A,\rho,\epsilon)$-upgrade time be
\begin{eqnarray}
\phi^{(A)}_{\rho,\epsilon}(\xi_{t_\rho})  = \min\left\{t'_\rho-t_\rho; \xi_{t'_\rho} \in \left(\bigcup_{i=\rho+1}^{n-2}E^{(A)}_{i}\right),t_\rho<\infty
 \right \}.
\end{eqnarray}
Similarly, the $(B,\rho,\epsilon)$-upgrade time $\phi^{(B)}_{\rho,\epsilon}(\xi_{t_\rho})$ can be defined. In the meantime, when a population $\xi$ has already $(A,\rho,\epsilon)$-taken over, the probability of generating at least one advanced LOIA in one generation, denoted by $u_{\rho,\epsilon}^{(A)}$, is given by
\begin{eqnarray*}
u_{\rho,\epsilon}^{(A)} = \min\left\{\sum_{\zeta \in \bigcup_{i=\rho+1}^{n-2}E^{(A)}_{i}} \mathbb{P}(\xi,\zeta); \xi \in E_{\rho,\epsilon N}  \right\},
\end{eqnarray*}
where $\mathbb{P}(\xi,\zeta)$ is the one-generation transition probability from population $\xi$ to population $\zeta$. The reciprocal of $u_{\rho,\epsilon}^{(A)}$ is an upper bound of the mean $(A,\rho,\epsilon)$-upgrade time, due to the property of geometric distribution \cite{Spiegel92book}. Similarly, we can define the upgrade probability $u_{\rho,\epsilon}^{(B)}$, and its reciprocal bounds the mean $(B,\rho,\epsilon)$-upgrade time from above.

\subsection{Bounding First Hitting Time}
As defined above, both the $(A,\rho,\epsilon)$-takeover time and $(B,\rho,\epsilon)$-takeover time describe the time sufficient for the EA to accumulate enough promising individuals for generating better individuals. Restricted on $\mathcal{S}^*$, the evolution of individuals towards the global optimum can be characterized as the so-called repeated takeover-upgrade process: First, the promising individuals are accumulated by the overall impact of the selection and mutation operators; When there are enough promising individuals in the population, the probability of generating better individuals becomes large enough, thus one or more individuals will soon upgrade to better individuals. The evolution of individuals restricted on $\mathcal{S}_0\cup(\mathcal{S}_1\setminus \mathcal{S}^*)$ can be characterized in a similar way, though the evolution will eventually lead to the local optimum (instead of the global optimum) if no individual belonging to $\mathcal{S}^*$ has been generated.

Formally, given $0\le \rho_1\le \rho_2\le n-2$ and generation index $t_{\rho_1}$ satisfying that the population $\xi_{t_{\rho_1}}$ belongs to $E_{\rho_1}^{(A)}$, define the mean first hitting time to $E_{\rho_2}^{(A)}$ of the $(N+N)$ EA starting from $E_{\rho_1}^{(A)}$ be
\begin{eqnarray}
\tau^{(A)}_{\rho_1,\rho_2}=\min\left\{t-t_{\rho_1};\xi_t\in \left(\bigcup_{i=\rho_2}^{n-2}E^{(A)}_{i}\right) ,t_{\rho_1}<\infty\right\}.
\end{eqnarray}
By drift analysis \cite{HeYao01} utilized in the proof of Proposition 1 in \cite{chen09unimodal}, we can easily combine the aforementioned repeated takeover-upgrade processes together and obtain the lemmas concerning the mean first hitting time to different population subset:
\begin{lemma}\label{lemma:unimodel_A}
On the \textsc{TrapZeros} problem, the mean first hitting time to $E_{\rho_2}^{(A)}$ of the $(N+N)$ EA starting from $E_{\rho_1}^{(A)}$ (where $0\le \rho_1\le \rho_2\le \ln^2n$ holds), conditional on that no individual belonging to $\mathcal{S}_0$ has ever been generated before the first time the EA reaches $E_{\rho_2}^{(A)}$, satisfies:
\begin{eqnarray}\label{Eq:lemma:A1}
\mathbb{E}\left[\tau^{(A)}_{\rho_1,\rho_2}\bigg|\bigwedge_{t=1}^{\tau^{(A)}_{\rho_1,\rho_2}}(\mathcal{S}_0\cap\xi_{t+1}=\emptyset)\right]= O\left[\sum_{k=\rho_1+1}^{\rho_2}\left(\bar{\eta}^{(A)}_{\epsilon}+1/u_{k,\epsilon}^{(A)}\right)\right],
\end{eqnarray}
where $\bar{\eta}^{(A)}_{\epsilon}$ is defined by Eq. \ref{eq:max_takeover}.

The mean first hitting time to $E_{\rho_2}^{(A)}$ of the $(N+N)$ EA starting from $E_{\rho_1}^{(A)}$ (where $\ln^2n\le \rho_1\le \rho_2\le n-2$ holds) satisfies:
\begin{eqnarray}\label{Eq:lemma:A2}
\mathbb{E}\left[\tau^{(A)}_{\rho_1,\rho_2}\right]= O\left[\sum_{k=\rho_1+1}^{\rho_2}\left(\bar{\eta}^{(A)}_{\epsilon}+1/u_{k,\epsilon}^{(A)}\right)\right];
\end{eqnarray}
\end{lemma}
Interested readers can refer to the appendix of this paper for the detailed proof of the above lemma. On the other hand, combining the takeover-upgrade processes for the LOIBs, we have the following lemma:
\begin{lemma}\label{lemma:unimodel_B}
On the \textsc{TrapZeros} problem, given the population size $N=\Omega(n/\ln n)$, the first hitting time to $E_{\rho_2}^{(B)}$ of the $(N+N)$ EA starting from $E_{\rho_1}^{(B)}$ (where $0\le \rho_1\le \rho_2\le n-2$ holds), conditional on that no individual belonging to $\mathcal{S}^*$ has ever been generated before the first time the EA reaches $E_{\rho_2}^{(B)}$, satisfies the following inequality with an overwhelming probability:
\begin{eqnarray}\label{eq:upper_bound}
\tau^{(B)}_{\rho_1,\rho_2}\le (\rho_2-\rho_1)\left(\hat{\eta}^{(B)}_{\epsilon} + \ln^3n\right),
\end{eqnarray}
where $\tau^{(B)}_{\rho_1,\rho_2}$ is the first hitting time to $E_{\rho_2}^{(B)}$ starting from $E_{\rho_1}^{(B)}$, $\hat{\eta}^{(B)}_{\epsilon}$ satisfies the following condition: $\forall i\in\{0,\dots,n-2\}, \forall\xi_{t_i}\in E_i^{(B)}: \mathbb{P}\left( \eta^{(B)}_{i,\epsilon}(\xi_{t_i})\le \hat{\eta}^{(B)}_{\epsilon} \right)\succ 1-1/SuperPoly(n)$, i.e., it is super-polynomially close to $1$.
\end{lemma}
The detailed proof is given in the appendix.

In this section, we have provided the analytical tools for the theoretical investigations of the $(N+N)$ EA. In the following parts of the paper, we will apply the tools to study the performance of the $(N+N)$ EA with different population sizes, so as to demonstrate the impact of population size on the performance of EA.

\section{$(N+N)$ EA with $N=1$}\label{sec:small}
In this section, we analyze the performance of $(1+1)$ EA on \textsc{TrapZeros}, where the $(1+1)$ EA can be considered as a degenerate case of the $(N+N)$ EA. By employing the aforementioned solvable rate as a measure, we obtain the following result:

\begin{theorem}
The first hitting time $\tau$ of the $(1+1)$ EA on \textsc{TrapZeros} is $O(n^2)$ with the probability of $\frac{1}{4}-O\big(\frac{\ln^2n}{n}\big)$. In other words, the solvable rate $\kappa$ of the $(1+1)$ EA on \textsc{TrapZeros} is at least $\frac{1}{4}-O\big(\frac{\ln^2n}{n}\big)$.
\end{theorem}
\textbf{\emph{Proof.}} After initialization, with a probability of $1/4$, the first and second bits of the initial individual both take the value $1$, i.e., the initial individual belongs to $\mathcal{S}_1$. At any generation, the probability that the EA generates an offspring belonging to $\mathcal{S}_0$ is $1/n^2$. Noting that the fitness of individual belonging to the schemata $\{(1,0,*,\dots,*)\}$ and $\{(0,1,*,\dots,*)\}$ is strictly smaller than that of any individual belonging to $\mathcal{S}_1$, the probability of avoiding finding any individual belonging to $\mathcal{S}_0$ at no later than the $t^{th}$ generation is at least $(1-1/n^2)^{t}/4$. Hence, for any $t\le (2en-1)\ln^2n$, the above probability is at least $(1-1/n^2)^{2en\ln^2n}/4>(1-4e\ln^2n/n)/4$.

Given the condition that the EA does not find any individual belonging to $\mathcal{S}_0$ before the $((2en-1)\ln^2n)^{th}$ generation, we then estimate the time sufficient for the EA to find some solutions belonging to the schema $\mathcal{S}^*$. The reason of concerning the schema $\mathcal{S}^*$ is that, once some individual belonging to $\mathcal{S}^*$ is found, the elitist selection operator will not accept any individual belonging to $\mathcal{S}_0$, which will eventually lead to the global optimum. It is easy to see that, at $t^{th}$ ($t\le (2en-1)\ln^2n$) generation of optimization, the EA takes the probability of no less than $(1/n)(1-1/n)^{n-3}\ge 1/(e^2n)$ to find an individual with better fitness, conditional on the event that the EA does not find any individual belonging to $\mathcal{S}_0$ before the $((2en-1)\ln^2n)^{th}$ generation. Hence, according to Chernoff bounds, with an extra probability of $1-e^{-\Theta(\ln^2n)}$ the EA can find some individual belonging to $\mathcal{S}^*$ no later than the $((2en-1)\ln^2n)^{th}$ generation.

Once the EA has found a solution belonging to $\mathcal{S}^*$, it will continue to find better solutions and eventually find the global optimum. In this phase, the EA takes the probability of no less than $(1/n)(1-1/n)^{n-3}\ge 1/(e^2n)$ to find an individual with better fitness, and in the worst case the Hamming distance between the individual and global optimum is $n-\ln^2n-2$. Hence, by Chernoff bounds, with an extra probability of $1-e^{-\Theta(n)}$, the EA can find the global optimum with extra $O(n^2)$ generations.

Combining the above propositions together, we have proven that the first hitting time $\tau$ of the $(1+1)$ EA on \textsc{TrapZeros} is $O(n^2)$ with a probability of $(1/4)(1-2e\ln^2n/n)(1-e^{-\Theta(\ln^2n)})(1-e^{-\Theta(n)})=1/4-O(\ln^2n/n)$.$\hfill\square$

The above theorem demonstrates that, when the population size is extremely small, the $(N+N)$ EA can find the global optimum of \textsc{TrapZeros} with a constant probability.

\section{$(N+N)$ EA with $N=O(\ln n)$} \label{sec:med}
In this section, the population size of the $(N+N)$ EA grows to $N=O(\ln n)$. The investigation begins with the following lemma concerning the number of LOIAs:
\begin{lemma}\label{lemma:N+N_cheby}
Let $X_t$ be the total number of LOIAs at the end of the $t^{th}$ generation. For the $(N+N)$ EA with truncation selection, we have
\begin{eqnarray*}
&&\mathbb{P}\bigg[X_{t+1}> (1+ch)X_t\Big| X_t<\frac{N}{1+ch}, \mathcal{S}_0\cap\xi_{t+1}=\emptyset \bigg]> 1-\frac{1-h}{(1-c)^2X_th},
\end{eqnarray*}
where $h$ is the probability that an old LOIA generates a new LOIA, $c\in(0,1)$ is a constant.
\end{lemma}
The proof of the above lemma is given in the appendix. Similar to Lemma \ref{lemma:N+N_cheby} focusing on LOIAs, the following lemma about LOIBs holds:
\begin{lemma}\label{lemma:N+N_cheby_B}
Let $Y_t$ be the total number of LOIBs at the end of the $t^{th}$ generation. For the $(N+N)$ EA with truncation selection, we have
\begin{eqnarray*}
&&\mathbb{P}\bigg[Y_{t+1}> (1+ch')Y_t\Big| Y_t<\frac{N}{1+ch'}, \mathcal{S}^*\cap\xi_{t+1}=\emptyset \bigg]> 1-\frac{1-h'}{(1-c)^2Y_th'},
\end{eqnarray*}
where $h'$ is the probability that an old LOIB generates a new LOIB, $c\in(0,1)$ is a constant.
\end{lemma}
For the sake of brevity, we do not offer the detailed proof of the above lemma. Interested readers can refer to the proof of Lemma \ref{lemma:N+N_cheby} for details.

By Lemma \ref{lemma:N+N_cheby}, we are able to bound the takeover time from above, which enables us to prove the following result:
\begin{theorem}\label{theorem:small_population}
Given the population size $N$ satisfying $N=O(\ln n)$ and $N=\omega(1)$, the first hitting time $\tau$ of the $(N+N)$ EA on \textsc{TrapZeros} is $O\big(\frac{n^2}{N}\big)$ with a probability of $1/Poly(n)$ \footnote{In this paper, $1/Poly(n)$ refers to some positive function (of the problem size $n$), whose reciprocal is bounded from above by a polynomial function of the problem size $n$.}. In other words, the solvable rate $\kappa$ of the $(N+N)$ EA on \textsc{TrapZeros} is at least $1/Poly(n)$, which is the reciprocal of some polynomial function of the problem size $n$.
\end{theorem}
\emph{\textbf{Proof.}} The proof requires two major steps which aim at proving the following propositions respectively:
\begin{enumerate}
\item[\ref{theorem:small_population}.1] The probabilities of the following two events are both $1/Poly(n)$: 1) The initial individuals all belong to $\mathcal{S}_1$; 2) The EA does not find any individual belonging to $\mathcal{S}_0$ within $n\ln^3n/N$ generations.
\item[\ref{theorem:small_population}.2] The maximal $(A,\rho,5/(5+c))$-takeover time ($\rho\in\{1,\dots,n-2\}$, $c\in(0,1)$ is a constant) of the $(N+N)$ EA on \textsc{TrapZeros}, conditional on the above two events, is upper bounded.
\end{enumerate}
Given the upper bound of the takeover times obtained when proving Proposition \ref{theorem:small_population}.2, we can utilize Lemma \ref{lemma:unimodel_A} to obtain a conditional expected first hitting time of the EA, which immediately leads to the theorem according to \ref{lemma:markov_inequality} (Markov inequality).
\subsubsection*{Proof of Proposition \ref{theorem:small_population}.1}
The proof begins with the discussions on Proposition \ref{theorem:small_population}.1. After the initialization of the EA, with the probability of $1/4^N$, the first and second bits of all initial individuals all take the value of $1$, i.e., the initial individuals all belong to $\mathcal{S}_1$. In the meantime, the probability for an individual (belonging to $\mathcal{S}_1$) to generate an offspring belonging to $\mathcal{S}_0$ is $1/n^2$. On the other hand, since the fitness of every individual belonging to the schemata $\{(1,0,*,\dots,*)\}$ or $\{(0,1,*,\dots,*)\}$ is strictly smaller than that of any individual belonging to $\mathcal{S}_1$, the truncation selection of the $(N+N)$ EA will always immediately eliminate all the newly generated individual belonging to the schemata $\{(1,0,*,\dots,*)\}$ and $\{(0,1,*,\dots,*)\}$. As a consequence of the above facts, the probability of avoiding finding any individual belonging to $\mathcal{S}_0$ before the $t^{th}$ generation is at least $(1-1/n^2)^{Nt}/4^N$. In other words, we have
\begin{eqnarray}\label{equation:event_prob}
\mathbb{P}\Bigg[\bigwedge_{t=1}^{n\ln^3n/N}(\mathcal{S}_0\cap\xi_{t}=\emptyset)\Bigg] \ge\frac{\big(1-\frac{1}{n^2}\big)^{n\ln^3n}}{4^N}>\frac{1-\frac{2\ln^3n}{n}}{4^N}.
\end{eqnarray}
So far we have proven Proposition \ref{theorem:small_population}.1.

\subsubsection*{Proof of Proposition \ref{theorem:small_population}.2}
Concerning Proposition \ref{theorem:small_population}.2, we focus on the $(A,\rho,\epsilon)$-takeover time, where $\rho\in\{1,\dots,n-2\}$ holds. Here we need to consider two cases according to the value of $\rho$: 1) $\rho\in\{1,\dots,\ln^2n\}$; 2) $\rho\in\{\ln^2n+1,\dots,n-2\}$.

Let us study the first case first, where $\rho\in\{1,\dots,\ln^2n\}$ holds. We note that at every generation the probability that an old LOIA generates a new LOIA is no smaller than $(1-p_m)^n=(1-1/n)^n> 1/5$ ($n\ge 2$), where $p_m=1/n$ is the mutation rate of the EA and $(1-p_m)^n$ is the probability that an old LOIA generates an offspring that is the same to itself. Hence, in the $(A,\rho,5/(5+c))$-takeover process (where we let $\epsilon=5/(5+c)$, and $c\in(0,1-\sqrt{\frac{4}{5}})$ is a positive constant) starting with $X_t=1$ (we consider the population with a unique LOIA at the beginning of the $(A,\rho,5/(5+c))$-takeover process in our worst-case analysis), the expected number of generations spent by the EA to accumulate $5$ LOIAs is less than $5/(1/5)=25$ generations.

On the other hand, for $X_t\in[5,5N/(5+c)]$, according to Lemma \ref{lemma:N+N_cheby}, we further obtain the following inequality
\begin{eqnarray}
\nonumber&&\mathbb{P}\bigg[X_{t+1}> (1+c/5)X_t\Big| X_t\in[5,5N/(5+c)],
\mathcal{S}_0\cap\xi_{t+1}=\emptyset \bigg]\\
\nonumber&>&\mathbb{P}\bigg[X_{t+1}> (1+ch)X_t\Big| X_t\in[5,5N/(5+c)], \mathcal{S}_0\cap\xi_{t+1}=\emptyset \bigg]> 1-\frac{1-h}{(1-c)^2X_th}\\
\label{equation:one_success_takeover}&\ge& 1-\frac{4}{5(1-c)^2}=\sigma,
\end{eqnarray}
where $c\in(0,1-\sqrt{\frac{4}{5}})$ is a constant, $h\ge (1-1/n)^n>1/5$ is the probability that an old LOIA generates a new LOIA, and $\sigma=1-4/(5(1-c)^2)<1$ is a positive constant. Now we can estimate upper bound for the takeover times, conditional on the event
\begin{eqnarray}
\label{equation:wedge_multi_event}\bigwedge_{t=1}^{n\ln^3n/N}(\mathcal{S}_0\cap\xi_{t}=\emptyset).
\end{eqnarray}
Here we call the event ``$X_{t+1}> (1+c/5)X_t$'' a success. Let $\hat{l}$ be the number of successes sufficient for the population to $(A,\rho,\epsilon)$-takeover, which can be estimated by
\begin{eqnarray}
\hat{l}=\min\left\{l;5\cdot\Big(1+\frac{c}{5}\Big)^l\ge\lceil\epsilon N\rceil\right\},
\end{eqnarray}
where $\epsilon = 5/(5+c)$. As a consequence, we obtain
\begin{eqnarray*}
\hat{l}\in\left(\frac{\ln5-\ln\lceil\epsilon N\rceil}{\ln\epsilon}-1,\frac{\ln5-\ln\lceil\epsilon N\rceil}{\ln\epsilon}+1\right).
\end{eqnarray*}
Hence, $\hat{l}=\Theta(\ln N)$ successes are sufficient for $(A,\rho,5/(5+c))$-takeover (in addition to an average of $25$ generations that are sufficient for accumulating 5 LOIAs). According to Eq. \ref{equation:one_success_takeover}, the expected number of generations that is sufficient for $(A,\rho,5/(5+c))$-takeover ($\rho\in\{1,\dots,\ln^2n\}$), conditional on the event described in Eq. \ref{equation:wedge_multi_event}, is at most $\hat{l}/\sigma+25$.
%By Chernoff bounds, the probability that the EA spends $2\hat{l}\ln n/\sigma$ generations (in addition to the $\delta$ generations required for the population to contain at least two LOIAs) to complete an $\epsilon$-takeover, conditional on Eq. \ref{equation:wedge_multi_event}, is $1-e^{-\hat{l}\ln n/(4\sigma)}$, which is an overwhelming probability. Given that the population size $N=O(\ln n)$, we further obtain that the probability that the EA spends $2\hat{l}\ln n/\sigma+\delta=O(\ln n\ln\ln n)$ generations to complete the $(A,\rho,\epsilon)$-takeover process (where $\rho\in\{1,\dots,\ln^2n\}$ holds), \emph{conditional on Eq. \ref{equation:wedge_multi_event}}, is $(1-e^{-\hat{l}\ln n/(4\sigma)})(1-(1-1/e)^{\ln n\cdot\ln\ln n})$, which is an overwhelming probability.

Similar to the case of $\rho\in\{1,\dots,\ln^2n\}$, we study the $(A,\rho,5/(5+c))$-takeover time for the case of $\rho\in\{\ln^2n+1,\dots,n-2\}$. The only difference between the two cases is that the former is carried out in the context of the condition that the individuals belonging to $\mathcal{S}_0$ have not been generated at every generation (which is summarized in Eq. \ref{equation:wedge_multi_event}) while the latter does not require such a condition. The obtained takeover time is similar: the expected number of generations sufficient for any $(A,\rho,5/(5+c))$-takeover ($\rho\in\{\ln^2n+1,\dots,n-2\}$) is at most $\hat{l}/\sigma+25$.

%The probability that the EA spends at most $2\hat{l}\ln n/\sigma+\delta=O(\ln n\ln\ln n)$ generations to complete the $(A,\rho,\epsilon)$-takeover process (where $\rho\in\{\ln^2n+1,\dots,n-2\}$ holds) is $(1-e^{-\hat{l}\ln n/(4\sigma)})(1-(1-1/e)^{\ln n\cdot\ln\ln n})$, which is an overwhelming probability.

In the above analysis, we have analyzed in detail the $(A,\rho,5/(5+c))$-takeover times for any $\rho\in\{1,\dots,n-2\}$, which completes the proof of Proposition \ref{theorem:small_population}.2.

\subsubsection*{Proof of Result}
The rest of the proof is to estimate the $(A,\rho,5/(5+c))$-upgrade times $1/u^{(A)}_{\rho,5/(5+c)}$ with respect to the $(A,\rho,5/(5+c))$-takeover times, and then follows the technique introduced Lemma \ref{lemma:unimodel_A}. Concerning the former point, we still consider two cases as we have done for estimating the $(A,\rho,5/(5+c))$-takeover times. For the first case, where $\rho\in\{1,\dots,\ln^2n\}$ holds, we estimate the $(A,\rho,5/(5+c))$-upgrade time $1/u^{(A)}_{\rho,5/(5+c)}$ under the condition described in Eq. \ref{equation:wedge_multi_event}: Once the population of the EA belongs to $E^{(A)}_{\rho,\lceil\epsilon N\rceil}\cup E^{(A)}_{\rho,\lceil\epsilon N\rceil+1}\cup\dots\cup E^{(A)}_{\rho,N}$, $u^{(A)}_{\rho,\epsilon}$ can be calculated as follows:
\begin{eqnarray*}
u_{\rho,\epsilon}^{(A)}= 1-\Bigg[1-\frac{1}{n}\bigg(1-\frac{1}{n}\bigg)^{\rho}\Bigg]^{\lceil\epsilon N\rceil}\ge1-\bigg(1-\frac{1}{e^2n}\bigg)^{\lceil\epsilon N\rceil}=1-\Bigg[\bigg(1-\frac{1}{e^2n}\bigg)^{e^2n}\Bigg]^{\frac{\lceil\epsilon N\rceil}{e^2n}},
\end{eqnarray*}
where $\epsilon=5/(5+c)$. Hence we have:
\begin{eqnarray*}
1/u_{\rho,\epsilon}^{(A)}\le\frac{1}{1-\left[\big(1-\frac{1}{e^2n}\big)^{e^2n}\right]^{\frac{\lceil\epsilon N\rceil}{e^2n}}}\le \frac{e^{\frac{\lceil\epsilon N\rceil}{e^2n}}}{e^{\frac{\lceil\epsilon N\rceil}{e^2n}}-1}\le 1+\frac{e^2n}{\lceil\epsilon N\rceil}.
\end{eqnarray*}
Similarly, for the second case $\rho\in\{\ln^2n+1,\dots,n-2\}$, we can also obtain the corresponding $(A,\rho,\epsilon)$-upgrade time:
\begin{eqnarray*}
1/u_{\rho,\epsilon}^{(A)}\le 1+\frac{e^2n}{\lceil\epsilon N\rceil}=O\Big(\frac{n}{N}\Big).
\end{eqnarray*}
The only difference between the two cases is that the proposition for the first case ($\rho\in\{1,\dots,\ln^2n\}$) holds when the condition described in Eq. \ref{equation:wedge_multi_event} holds, while the second case ($\rho\in\{\ln^2n+1,\dots,n-2\}$) does not require such a condition.

As a consequence, by Lemma \ref{lemma:unimodel_A}, we can estimate the upper bound of the expected first hitting time (to the population subset $E_{\ln^2n+1}^{(A)}$) of the EA, conditional on the event described in Eq. \ref{equation:wedge_multi_event}. Formally, let $\tau_{\mathcal{S}^*}$ be the first hitting time (to the population subset $E_{\ln^2n+1}^{(A)}$), $\bar{\tau}_{\mathcal{S}^*}$ be the above conditional expectation. According to Lemma \ref{lemma:unimodel_A} derived from the technique presented in \cite{chen09unimodal}, the asymptotic order of such a conditional expectation is no larger than $\ln^2n(\bar{\eta}_{\epsilon}+\max_{\rho\le \ln^2n}\{1/u^{(A)}_{\rho,\epsilon}\}) \le\ln^2n \left(\hat{l}/\sigma+25+O(n/N)\right)=\ln^2n\left(O(\ln\ln n)+O(n/N)\right)=O(\ln^2 n\ln\ln n+n\ln^2 n/N)$. According to Lemma \ref{lemma:markov_inequality} (Markov inequality), under the condition described in Eq. \ref{equation:wedge_multi_event}, with the probability of at least $1/2$, the first hitting time to the population subset $E_{\ln^2n+1}^{(A)}$ (i.e., $\tau_{\mathcal{S}^*}$) is less than $2\bar{\tau}_{\mathcal{S}^*}=\omega(n\ln^3n/N)$. Combining this result with the proof of Proposition \ref{theorem:small_population}.1 (which estimates the probability that the condition described in Eq. \ref{equation:wedge_multi_event} holds), we know that with a probability of $(1-2\ln^3n/n)\cdot(1/2)/4^N$, the EA has found an individual belonging to $\mathcal{S}^*$ within $2\bar{\tau}_{\mathcal{S}^*}=O(\ln^2 n\ln\ln n+n\ln^2 n/N)$ generations.

Afterwards, following the same technique, we further bound from above the expected first hitting time to $E_{n-2}^{(A)}$ (the population subset consisting of all populations containing the global optimum), with the starting point (initial population) $\xi_1\in\left(\bigcup_{j=\ln^2n+1}^{n-2}E^{(A)}_j\right) $. Formally, given the above starting point of the EA, let $\tau'$ be the first hitting time to $E_{n-2}^{(A)}$, $\bar{\tau}'$ be the expectation of $\tau'$. According to Lemma \ref{lemma:unimodel_A}, the asymptotic order of $\bar{\tau}'$ is no larger than $(n-\ln^2n-2)(\bar{\eta}_{\epsilon}+1/u_{\rho,\epsilon}^{(A)}) =(n-\ln^2n-2) \left(O(\ln\ln n)+O(n/N)\right)=O(n\ln\ln n+n^2/N)$. By Markov inequality, with a probability of at least $1/2$, the first hitting time $\tau'$ is less than $2\bar{\tau}'=O(n\ln\ln n+n^2/N)$.

Combining the above results together, we obtain that the first hitting time of the $(N+N)$ EA (i.e., $\tau_{\mathcal{S}^*}+\tau'$) is upper bounded by $2\bar{\tau}_{\mathcal{S}^*}+2\bar{\tau}'=O(n^2/N)$ with a probability of $(1-2\ln^3n/n)\cdot(1/2)\cdot(1/2)/4^N=(1-2\ln^3n/n)/4^{N+1}$. Noting that the population size satisfies $N=O(\ln n)$, the above probability is $1/Poly(n)$. Hence, we have proven the whole theorem. $\hfill\square$

\section{$(N+N)$ EA with $N=\Omega(n/\ln n)$}\label{sec:large}
In this section, we further increase the population size, and study the performance of the EA on the basis of our previous results. We have the following result:
\begin{theorem}\label{th:large_pop} Given the polynomial population size $N$ satisfying $N=\Omega(n/\ln n)$, the first hitting time $\tau$
of the $(N+N)$ EA on \textsc{TrapZeros} is super-polynomial with an overwhelming probability. In other words, the solvable rate $\kappa$ of the $(N+N)$ EA on \textsc{TrapZeros} is super-polynomially close to $0$.
\end{theorem}
\emph{\textbf{Proof.}} The proof of Theorem \ref{th:large_pop} contains several steps, in which we focus on different propositions required for proving the whole theorem:
\begin{enumerate}
\item[\ref{th:large_pop}.1] After initialization, the probability that all the individuals in the population contain no more than $3\ln^2n/4$ 1-bits among the leading $\ln^2n+2$ bits is super-polynomially close to $1$.
\item[\ref{th:large_pop}.2] With an overwhelming probability, the EA cannot find any individual belonging to $\mathcal{S}^*$ before the $\eta^{th}$ generation, where $\eta=\ln N\cdot\ln\ln n$. With an overwhelming probability, there are still at least $\ln^2 n/16$ 0-bits between the $(\ln^2 n/8)^{th}$ and $(\ln^2n+3)^{th}$ bits of each individual at the end of the $\eta^{th}$ generation.

\item[\ref{th:large_pop}.3] No later than the $\eta^{th}$ generation, the population will be taken over by the individuals belonging to $\mathcal{S}_0$ with an overwhelming probability.

\item[\ref{th:large_pop}.4] After the $\eta^{th}$ generations, the probability that an individual belonging to $\mathcal{S}^*$ is generated via direct mutation from an individual belonging to $\mathcal{S}_0$ is super-polynomially close to $0$.
\end{enumerate}
\subsubsection*{Proof of Proposition \ref{th:large_pop}.1}

The polynomial population size $N=\Omega(n/\ln n)$ implies that the initial population contains one or more individuals belonging to $\mathcal{S}_0$ with the probability of $1-(1-1/4)^{N}=1-(3/4)^{\Omega(n/\ln n)}$, which is an overwhelming probability. Meanwhile, the probability of generating one or more individuals belonging to $\mathcal{S}^*$ is $1-(1-(1/2)^{\ln^2n+2})^N\approx N/2^{\ln^2n+2}$, which is super-polynomially close to $0$ due to that $N$ is a polynomial function of the problem size $n$. Hence, after initialization the best individual in the population belongs to $\mathcal{S}_0$ with an overwhelming probability. On the other hand, concerning the number of 1-bits among the leading $\ln^2n+2$ bits for every initial individual, we apply Chernoff bounds, and obtain that the probability of having no more than $3\ln^2n/4$ 1-bits is $1-e^{-\ln^2n/32}$, which is an overwhelming probability. As a consequence, after initialization, the probability that all the individuals in the population contain no more than $3\ln^2n/4$ 1-bits among the leading $\ln^2n+2$ bits is also an overwhelming one.

\subsubsection*{Proof of Proposition \ref{th:large_pop}.2}
As mentioned, we define $\eta=\ln N\cdot\ln\ln n=O(\ln n\cdot\ln\ln n)$ ($N=\Omega(n/\ln n)$), we now prove that within $\eta$ generations, the probability of finding individuals belonging to $\mathcal{S}^*$ is super-polynomially close to $0$. Denote by $\mathcal{C}$ the event ``the first 0-bit among the leading $\ln^2n+2$ bits of an individual is flipped, while the 1-bits before the flipped 0-bit (i.e., the leading 1-bits) have not been flipped''.
%Here we carry out a best-case analysis by
%assuming that within $\eta$ generations, every individual takes the
%probability of $1/n$ to have a success. Moreover,
Optimistically, we assume that event $\mathcal{C}$ always happens before the $\eta^{th}$ generation. Under the circumstance, with the overwhelming probability $1-\Theta(1/n^{\ln\ln n})$, at most $\ln\ln n$ 0-bits among the leading $\ln^2n+2$ bits of an individual can be flipped simultaneously at each generation. Given a polynomial population size $N$, the above proposition implies that at every generation, the maximal number of 1-bits among the leading $\ln^2n+2$ bits of each individual in the population can increase by at most $\ln\ln n$ with an overwhelming probability. Combining the above fact with the result presented in Proposition \ref{th:large_pop}.1, we know that in order to find an individual belonging to $\mathcal{S}^*$, we need at least $(\ln^2n+2-3\ln^2n/4)/(\ln\ln n)>\ln^2n/(4\ln\ln n)$ generations with an overwhelming probability, which implies that $\eta=o(\ln^2n/(4\ln\ln n))$ generations are not enough for the EA to generate an individual belonging to $\mathcal{S}^*$ with an overwhelming probability.

In the meantime, the fact that (with an overwhelming probability) at most $\ln\ln n$ 0-bits among the leading $\ln^2n+2$ bits of an individual can be flipped simultaneously at each generation also implies that (with an overwhelming probability) there are still at least $7\ln^2n/8+2-3\ln^2n/4-O(\eta\cdot(\ln\ln n))>\ln^2 n/16$ 0-bits between the $(\ln^2 n/8)^{th}$ and $(\ln^2n+3)^{th}$ bits of each individual at the end of the $\eta^{th}$ generation.

\subsubsection*{Proof Sketch of Proposition \ref{th:large_pop}.3}
Next, we need to prove that within $\eta$ generations, the population will be taken over by the individuals belonging to $\mathcal{S}_0$ (i.e., $(B,0,1)$-takeover) with an overwhelming probability. This proposition can be proven by a technique similar to the one utilized in the proof of Theorem \ref{theorem:small_population}, i.e., by applying Chebyshev inequality (Lemma \ref{lemma:N+N_cheby_B}) for establishing the probability of ``a success''. More specifically, let $Y_t$ be the number of LOIBs at the $t^{th}$ generation. For $Y_t\in[5,5N/(5+c)]$, we define the event ``$Y_{t+1}> (1+c/5)Y_t$'' be a ``success''. According to Lemma \ref{lemma:N+N_cheby_B}, we obtain
\begin{eqnarray}
\nonumber&&\mathbb{P}\bigg[Y_{t+1}> (1+c/5)Y_t\Big| Y_t\in[5,5N/(5+c)],
\mathcal{S}^*\cap\xi_{t+1}=\emptyset \bigg]\\
\nonumber&>&\mathbb{P}\bigg[Y_{t+1}> (1+ch')Y_t\Big| Y_t\in[5,5N/(5+c)], \mathcal{S}^*\cap\xi_{t+1}=\emptyset \bigg]> 1-\frac{1-h'}{(1-c)^2Y_th'}\\
\nonumber&\ge& 1-\frac{4}{5(1-c)^2}=\sigma',
\end{eqnarray}
where $c\in(0,1-\sqrt{\frac{4}{5}})$ is a constant, $h'\ge (1-1/n)^n>1/5$ is the probability that an old LOIB generates a new LOIB, and $\sigma'=1-4/(5(1-c)^2)<1$ is a positive constant. For the sake of brevity, we omit the mathematical details of discussing the number of successes sufficient for $(B,0,1)$-takeover. One can refer to the proof of Proposition \ref{theorem:small_population}.2 for more details. Next we prove that the number of generations sufficient for $(B,0,1)$-takeover is no larger than $\eta$ with an overwhelming probability.

Assume that we have already proven that the number of successes sufficient for $(B,0,1)$-takeover (using Lemma \ref{lemma:N+N_cheby_B} and the proof idea of Proposition \ref{theorem:small_population}.2), denoted by $\hat{l}'$, satisfies $\hat{l}'\le \ln N/\ln(1/\epsilon)+3$ (where $\epsilon=5/(5+c)$, $c$ is a constant), and the probability of achieving a success at each generation (belonging to the $(B,0,1)$-takeover process) is no less than a constant $\sigma'\in(0,1)$. For the $(B,0,1)$-takeover process, let $\mbox{suc}(T)$ be the number of successes happened among $T$ generations. When the number of successes is smaller than $\hat{l}'$  \footnote{When the number of successes $\mbox{suc}(T)$ reaches $\hat{l}'$, the population has been $(B,0,1)$-taken over. Afterwards, if no individual belonging to $\mathcal{S}^*$ has been generated (the corresponding duration has been investigated by Proposition \ref{th:large_pop}.2), the event ``success'' still makes sense except that the population has entered the $(B,\rho,1)$-takeover process, where $\rho>0$.}, the evolution at each generation can be considered as a Bernoulli trial: the probability of having a success is at least $\sigma'$. According to Chernoff bounds, the probability that among $T$ generations the number of successes $\mbox{suc}(T)$ is no smaller than $(1-\delta)\cdot\mathbb{E}[\mbox{suc}(T)]\ge (1-\delta)\cdot (T\sigma')$ is at least $1-e^{-T\sigma'\delta^2/2}$, where $\delta\in(0,1)$ is a constant. By setting $T=\sqrt{\ln\ln n}\cdot \hat{l}'/((1-\delta)\sigma')=\Theta(\ln N\cdot\sqrt{\ln\ln n})$, we obtain that, the probability that $T=\Theta(\ln N\cdot \sqrt{\ln\ln n})$ generations are sufficient for obtaining more than $\hat{l}'=\Theta(\ln N)$ successes is at least $1-e^{-T\sigma'\delta^2/2}=1-e^{-\Theta(\ln N\cdot\sqrt{\ln\ln n})\sigma'\delta^2/2}$, which is an overwhelming probability. By summarizing the above discussions, we know that with an overwhelming probability it takes at most $\Theta(\ln N\cdot \sqrt{\ln\ln n})$ generations for the population to be $(B,0,1)$-taken over. Since $\eta=\ln N\cdot\ln\ln n=\omega(\ln N\cdot \sqrt{\ln\ln n})$, we have proven Proposition \ref{th:large_pop}.3.

Here, it is worth noting that the above proposition, along with the truncation selection that kills the worst $N$ individuals among the overall $2N$ parents and offsprings, is getting very close to the final conclusion of the theorem. More precisely, once the population has been $(B,\rho,1)$-taken over by the individuals belonging to $\mathcal{S}_0$ (for any $\rho=0,\dots,n-2$), the truncation selection operator will no longer accept offsprings that belonging to neither $\mathcal{S}^*$ nor $\mathcal{S}_0$, since these offsprings have lower fitness than all $N$ parents belonging to $\mathcal{S}_0$. Hence, the only way to generate individuals belonging to $\mathcal{S}^*$ is via the direct mutations of those parent individuals belonging to $\mathcal{S}_0$. Next, we show that such a probability is super-polynomially small.

\subsubsection*{Proof of Proposition \ref{th:large_pop}.4}
Optimistically, here we can assume that the bits between the $(\ln^2 n/8)^{th}$ and $(\ln^2n+3)^{th}$ bits of each individual have become ``free-riders'' \cite{Droste02} when the population has just been $(B,\rho,1)$-taken over (where $\rho\in \{0,\dots,\ln^2n/8-3 \}$), i.e., the values of these bits are not influenced by selection pressure, and only genetic drift is considered (though some of them are very likely to be influenced by selection pressure which tends to preserve 0-bits). By dividing the evolutions after the $\eta^{th}$ generation into two stages, we need to prove the following propositions:
\begin{enumerate}
\item[\ref{th:large_pop}.4.1] No later than the $(\eta+\alpha)^{th}$ generation, the population has been $(B,\rho,1)$-taken over by individuals
with no less than $\ln^2 n/8$ consecutive leading 0-bits, where $\alpha=\ln^5n$.

\item[\ref{th:large_pop}.4.2]  After the $\eta^{th}$ generation but before the population has been taken over by individuals with no less than $\ln^2 n/8$ consecutive
leading 0-bits\footnote{$(B,\ln^2 n/8-2,1)$-takeover, happened no later than the $(\eta+\alpha)^{th}$ generation with an overwhelming probability.}, with an overwhelming probability at least $\ln^2 n/32$ free-riders between the $(\ln^2 n/8)^{th}$ and $(\ln^2n+3)^{th}$ bits of each individual (in the population) will take the value of $0$.
\end{enumerate}

By Lemma \ref{lemma:unimodel_B}, we can study the $(B,\rho,\epsilon)$-takeover processes ($\rho\le\ln^2 n/8-2$) and the corresponding $(B,\rho,\epsilon)$-upgrade probability, and prove that the number of generations sufficient for the population to be $(B,\ln^2 n/8-2,1)$-taken over by individuals with more than $\ln^2 n/8$ consecutive leading 0-bits is bounded from above by $(\ln^2 n/8)(\hat{\eta}^{(B)}_{\epsilon}+\max_{\rho\le \ln^2 n/8-2}\{1/u_{\rho,1}^{(B)}\})<(\ln^2 n/8)(\hat{\eta}^{(B)}_{\epsilon}+\ln^3n)$ with an overwhelming probability, where $\hat{\eta}^{(B)}_{\epsilon}=o(\ln N\cdot\ln\ln n)=o(\ln n\cdot\ln\ln n)$. The proof follows the proof idea of Proposition \ref{th:large_pop}.3 to study the takeover time $\hat{\eta}^{(B)}_{\epsilon}$, and then applies Lemma \ref{lemma:unimodel_B} directly. For the sake of brevity, here we do not provide the details. Moreover, the above result, along with the condition $N=\Omega(n/\ln n)$, implies that $\eta+(\ln^2 n/8)(\hat{\eta}^{(B)}_{\epsilon}+\ln^3n )<\eta+\ln^5n=\eta+\alpha$. Hence, we can reach Proposition \ref{th:large_pop}.4.1.

On the other hand, recall that when the population has just been taken over by the individuals belonging to $\mathcal{S}_0$ (i.e., $(B,\ln^2 n/8-2,1)$-takeover), with an overwhelming probability there are still at least $\ln^2 n/16$ free-riders, between the $(\ln^2 n/8)^{th}$ and $(\ln^2n+3)^{th}$ bits of each individual, taking the value of $0$. Within $\ln^5n$ generations, each of the free-riders will receive at most $\alpha=\ln^5 n$ mutations. Given any individual at the $\eta^{th}$ generation, there are at least $\ln^2 n/16$ free-riders (between its $(\ln^2 n/8)^{th}$ and $(\ln^2n+3)^{th}$ bits) taking the value of $0$. For each of those free-riders, the probability that its value does not change within the $\alpha$ mutations is at least $(1-1/n)^\alpha\approx 1-O(\ln^5 n/n)$. According to Chernoff bounds, among the aforementioned $\ln^2 n/16$ free-riders of each individual, with the overwhelming probability of $1-e^{-\Theta(\ln^2 n)}=1-n^{-\Theta(\ln n)}$, there are still $\ln^2 n/32$ free-riders consistently taking the value of $0$ between the $\eta^{th}$ generation and the $(\eta+\alpha)^{th}$ generation. As a consequence, the fact that the population size $N$ is a polynomial function of the problem size $n$ yields Proposition \ref{th:large_pop}.4.2.

By summarizing the above discussions on the number of free riders, we know that before the population has been taken over by individuals with no less than $\ln^2 n/8$ consecutive leading 0-bits (but after the $\eta^{th}$ generation), the probability for each individual to find an offspring belonging to $\mathcal{S}^*$ is at most $1/n^{\ln^2n/32}$. On the other hand, after the population has been taken over by individuals with no less than $\ln^2 n/8$ consecutive leading 0-bits, the probability for each individual to find an offspring belonging to $\mathcal{S}^*$ is $1/n^{\ln^2n/8}$, since at that time each individual in the population will contain at least $\ln^2 n/8$ consecutive leading 0-bits. Further, since the population size $N$ is polynomial, the probability that some individual in the population generates an offspring belonging to $\mathcal{S}^*$ is still super-polynomially close to $0$.

Combining the results presented in Propositions \ref{th:large_pop}.1, \ref{th:large_pop}.2, \ref{th:large_pop}.3 and \ref{th:large_pop}.4 together, we have proven the theorem. $\hfill\square$
%\begin{theorem}
%Given the population size $N=O\big(\frac{n^2}{\ln^3 n}\big)$ and
%Initialization II, the first hitting time $\tau$ of the $(N+N)$ EA
%on \textsc{TrapZeros} is $O\big(\frac{n^2}{N}\big)$ with a
%probability of $1/Poly(n)$.
%\end{theorem}
%
%\begin{theorem}
%Given the population size $N=\omega\big(\frac{n^2}{\ln^3 n}\big)$
%and Initialization II, the first hitting time $\tau$ of the $(N+N)$
%EA on \textsc{TrapZeros} is exponential with an overwhelming
%probability.
%\end{theorem}
\section{Discussion}\label{sec:discussion}
So far we have seen three analytical results concerning the performance of a population-based EA with different population sizes. It is shown that with the increase of population size, the solvable rate of the $(N+N)$ EA will drop to an extremely low level on the \textsc{TrapZeros} problem. Although the study in this paper is carried out on a specific problem using a specific type of EAs, it has much wider implications. The analytical results presented in this paper actually demonstrate an interesting problem characteristic under which the population-based EAs may perform poorly: when a problem has an attraction basin leading to some local optimum, and the individuals at this basin are with relatively high fitness than most individuals, a large population may not be useful and even becomes harmful, since it will lead to a large probability of finding individuals at the local basin. The resultant takeover process at the mistaken basin will quickly eliminates other promising individuals that may lead to the global optimum. After that, only large step sizes can help to find promising individuals again, resulting in a long runtime of the EA due to the small probability of getting close to the global optimum.

The weakness of the population-based EAs without recombination on the above problem characteristic, shown in this paper, can partially be tackled by employing larger step sizes. For example, if some appropriate recombination strategy, which can generate large step sizes, is employed, the EA can probably provide much larger step sizes in comparison with bitwise mutation (adopting the commonly used mutation rate $1/n$). Moreover, some adaptive/self-adaptive mutation schemes might also be helpful to cope with the situation, since they can provide large step sizes in exploring the correct attraction basin, and small large sizes in exploiting the correct basin. As a consequence, even if the whole population has been trapped in a mistaken basin, it is still possible to find promising individuals in other basins of attraction. The related investigations will be left as our future work.

\section{Conclusion}\label{sec:Conclusion}
In this paper, we have investigated the performance of an $(N+N)$ EA with different population sizes on a multimodal problem, namely \textsc{TrapZeros}. The theoretical results have revealed a problem characteristic that may lead to poor performances of population-based EAs, as mentioned in the last section. This is the first time that the influence of population size on an $(N+N)$ EA is analyzed theoretically. In addition, the proposed solvable rate, which is an intrinsic feature extracted from the probability distribution of the first hitting time, offers an alternative choice for measuring the performance of EA.

Deriving from a recently-developed approach for analyzing EAs on unimodal problems \cite{chen09unimodal} and following the well-known building block hypothesis, the utilized takeover-upgrade technique is capable of characterizing the evolution within a single basin of attraction as repeated takeover-upgrade processes that accumulate enough promising individuals and then generate better individuals by the accumulated individuals. In this paper, the successful application of this technique in modeling population-based EAs on a multimodal problem has shed some light on analyzing more complicated population-based EAs using similar techniques. To utilize such techniques, an elaborate procedure, as shown in the proof of Theorem \ref{theorem:small_population}, is required to estimate the takeover times, which is directly related to the selection operator. In the future, we will further study the EAs with different parent and offspring sizes, i.e., the $(\mu+\lambda)$ EAs. Moreover, we will combine the techniques with other state-of-the-art analytical tools, so as to gain more insight into the impact of other operators (e.g., recombination) and the corresponding parameter settings on the performance of EA.

\section*{Appendix}
\subsection*{Drift Analysis}
Drift analysis is a well-known technique for studying the time complexity of EAs \cite{HeYao01,HeYao04}. Formally, Let $x^*$ be the unique optimum of the objective function, let $V(X)$ be the function that measures the distance between the population $X$ and the optimum $x^*$, then the one step mean drift at the $t^{th}$ generation of the EA, denoted by $\Delta(X,t)$, is given by
\begin{eqnarray*}
\Delta(X,t)&=&\mathbb{E}[V(\xi_t)-V(\xi_{t+1})\mid \xi_t=X]\\
&=&\sum_{Y\in E}\left[V(X)-V(Y)\right] \mathbb{P}(X,Y;t),
\end{eqnarray*}
where $E$ is the whole population set. If no self-adaptive strategy is utilized in the EA, then the transition probability is not time dependent:
\begin{eqnarray*}
 \mathbb{P}(X,Y;t)=\mathbb{P}(X,Y).
\end{eqnarray*}
Under this circumstance, we simply use the notation $\Delta(X)$ to represent the one step mean drift. Meanwhile, if
\begin{eqnarray*}
 \Delta(X)=\mathbb{E}[V(\xi_t)-V(\xi_{t+1})\mid \xi_t=X]\ge 0
\end{eqnarray*}
 holds for $t=0,1,\dots$, then $\{V(\xi_t):t=0,1,\dots \}$ is called a super-martingale. According to He and Yao \cite{HeYao01,HeYao04}, once we can estimate the lower bound of the one step mean drift, then we can get the upper bound of the mean first hitting time by the following lemma:
\begin{lemma}[Drift Theorem\cite{HeYao01}]\label{driftcondition}
Let $\{V(\xi_t):t=0,1,\dots \}$ be a super-martingale describing an EA, if for any time $t=0,1,2,\dots$, if $V(\xi_t)>0$ and
\begin{eqnarray*}
\mathbb{E}[V(\xi_t)-V(\xi_{t+1})\mid \xi_t]\ge c_l>0,
\end{eqnarray*}
then the mean first hitting time satisfies
\begin{eqnarray*}
\mathbb{E}[\tau\mid \xi_0]\le \frac{V(\xi_1)}{c_l},
\end{eqnarray*}
where $\xi_0$ is the initial population of the EA.
\end{lemma}
One of the key steps when applying drift analysis is to specify an appropriate distance function $V(\cdot)$. The following lemma tells us that when the expected first hitting time of a homogeneous absorbing Markov chain, conditional on the starting state $X$, is defined as the distance function for the population state $X$, the one step mean drift equals $1$:
\begin{lemma}[\cite{HeYao04}]\label{distance1}Let $L:\{L_t, t=0,1,\dots\}$ be a homogeneous absorbing Markov chain defined on the space $M$, $H\subset M$ be its absorbing subspace. For $L$, its first hitting time to $H$, which is formally defined by
\begin{eqnarray*}
  \tau = \min\{t \ge 0; L_t \in H\},
\end{eqnarray*}
satisfies
\begin{eqnarray*}
\left\{
\begin{array}{llll}
\mathbb{E}[\tau|L_0=X]=0,\quad\quad\quad\quad\quad\quad\quad\quad\quad\quad\quad\quad\quad\mbox{ } X\in H;&\\
\mathbb{E}[\tau|L_0=X]-\sum_{Y\in M}\mathbb{P}(X,Y)\mathbb{E}[\tau|L_0=Y]=1, X\notin H,&
  \end{array}
  \right.
\end{eqnarray*}
where $\mathbb{E}[\tau|L_0=X]$ is the expected first hitting time of the absorbing Markov chain $L$ starting with the initial state $L_0=X$.
\end{lemma}
The proof of Lemma \ref{lemma:unimodel_A} will utilize the above lemmas.

\subsection*{Proof of Lemma \ref{lemma:unimodel_A}}
We offer the detailed proof for Eq. \ref{Eq:lemma:A1} in Lemma \ref{lemma:unimodel_A}. Before bounding the mean of the first hitting time $\tau^{(A)}_{\rho_1,\rho_2}$, we have to study the $(A,k,\epsilon)$-takeover process ($\rho_1\le k\le \rho_2\le\ln^2n$) by restricting the Markov chain on $E^{(A)}_k$ under the condition ``{\em no individual belonging to $\mathcal{S}_0$ has ever been generated before the first time the EA reaches $E_{\rho_2}^{(A)}$}''. Formally, the above condition is represented by
\begin{eqnarray*}
\bigwedge_{t=1}^{\tau^{(A)}_{\rho_1,\rho_2}}(\mathcal{S}_0\cap\xi_{t+1}=\emptyset).
\end{eqnarray*}
The rest of the proof will be presented in the context of the above condition (for the sake of brevity we omit it when presenting mathematical details).

The analysis follows a pessimistic style, which is quite similar to the proof of Proposition 1 in \cite{chen09unimodal}. If the number of the LOIAs of a population $Z$ is smaller than $\epsilon N$, we ignore all potential emergences of advanced LOIAs. Once some advanced LOIA is generated before the number of the LOIAs reaches $\epsilon N$, we assume pessimistically that each advanced LOIA is only a LOIA, i.e., it is replaced by some LOIA that is randomly selected from the LOIAs in the current population. In response to this step, the population $Z$ is transformed to $Y$ $(Y\in E^{(A)}_{k,i})$. Afterwards, according to the number of LOIAs, we consider two different cases: $\lceil\epsilon N\rceil\le i\le N$ and $1\le i<\lceil\epsilon N\rceil$, and use the notation ``$\to_k$'' to represent the mapping from $\bigcup_{j=k+1}^{n-2}E^{(A)}_j$ to $E^{(A)}_k$:
\begin{itemize}
\item{If $\lceil\epsilon N\rceil\le i\le N$}, then $Z\to_k Y$. (An advanced LOIA is transformed to a LOIA.)
\item{If $1\le i<\lceil\epsilon N\rceil$}, then $Z\to_k Y'$, where $Y'$ is obtained by
directly replacing the best $\lceil\epsilon N\rceil$ individuals of $Y$ with $\lceil\epsilon N\rceil$ randomly selected LOIAs, $Y'\in E^{(A)}_{k,\lceil\epsilon N\rceil}$ (We ignore the advanced LOIAs but consider that the population has $(A,k,\epsilon)$-taken over by $\lceil\epsilon N\rceil$ LOIAs).
\end{itemize}
The aim of the above transformations is to restrict the whole $(A,k,\epsilon)$-takeover process on the subspace $E^{(A)}_k$. The consequence is that we can utilize an auxiliary homogeneous absorbing Markov chain ($\zeta_t^{(k)}, t=0,1,\dots$) on $E^{(A)}_k$ to study the whole $(A,k,\epsilon)$-takeover process. The transition probabilities of the auxiliary Markov chain are given by {\small
\begin{eqnarray}
\label{equation:transition_probabilties} \bar{\mathbb{P}}(X,Y)=\left\{
\begin{array}{llll}
 \mathbb{P}(X,Y), &X\neq Y, \quad X \notin \bigcup_{i=\lceil\epsilon N\rceil}^{N} E^{(A)}_{k,i} , Y\notin \bigcup_{i=\lceil\epsilon N\rceil}^{N}
 E^{(A)}_{k,i};\\
 \mathbb{P}(X,Y)+\sum_{Z \in \bigcup_{j=k+1}^{n-2}E^{(A)}_j, Z\to_k Y} \mathbb{P}(X,Z), &X\neq Y,  \quad X\notin
 \bigcup_{i=\lceil\epsilon N\rceil}^{N} E^{(A)}_{k,i},  Y\in \bigcup_{i=\lceil\epsilon N\rceil}^{N} E^{(A)}_{k,i}; \\
 0, &X\neq Y,  \quad X\in \bigcup_{i=\lceil\epsilon N\rceil}^{N} E^{(A)}_{k,i}, Y\in \bigcup_{i=\lceil\epsilon N\rceil}^{N} E^{(A)}_{k,i};\\
0 , &X\neq Y,  \quad X\in \bigcup_{i=\lceil\epsilon N\rceil}^{N} E^{(A)}_{k,i},
Y\notin \bigcup_{i=\lceil\epsilon N\rceil}^{N} E^{(A)}_{k,i};\\
1-\sum_{Y \neq X} \bar{\mathbb{P}}(X,Y) , &X= Y,
  \end{array}
 \right.
\end{eqnarray}}
where $X,Y\in E^{(A)}_k$. According to the definitions of transition probabilities presented in Eq. \ref{equation:transition_probabilties}, we have
\begin{eqnarray*}\sum_{Y \in E^{(A)}_k} \bar{\mathbb{P}}(X,Y)=1.\end{eqnarray*}

Obviously, $\left(\bigcup_{i=\lceil\epsilon N\rceil}^{N} E^{(A)}_{k,i}\right)\cup \left(\bigcup_{j=k+1}^{n-2}E^{(A)}_j\right)$ is absorbing in $\zeta^{(k)}$. The other subspaces are $E^{(A)}_{k,1},\dots,$ $E^{(A)}_{k,\lceil\epsilon N\rceil-1}$. According to Lemma~\ref{distance1}, we know that for $X\in \bigcup_{i=1}^{\lceil\epsilon N\rceil-1} E^{(A)}_{k,i}$,
\begin{eqnarray} \label{m}
 \sum_{Y \in E^{(A)}_k} \left[\bar{\eta}_{k,\epsilon}^{(A)}(X)-\bar{\eta}_{k,\epsilon}^{(A)}(Y)\right] \bar{\mathbb{P}}(X,Y) = 1.
\end{eqnarray}
%According to Definition \ref{fht}, starting from the initial population $\xi_1$, the first hitting time to the subspace $E^{(A)}_{n-2}$, is given by
%\begin{eqnarray*}
%  \tau = \min\{t \ge 0; \xi_t \in E^{(A)}_{n-2}\}.
%\end{eqnarray*}

On the other hand, let $\xi_{t_k}$ be the population of the $t_k^{th}$ generation.
 For $\xi_{t_k}\in E^{(A)}_k$, we define its first hitting time to the population set
 $\bigcup_{j=k+1}^{n-2}E^{(A)}_j$, starting from $\xi_{t_k}$:
\begin{eqnarray*}
\tau^{(A)}_{k,k+1}=\min\left\{t'_k -t_k\ge 0; \xi_{t'_k} \in \bigcup_{j=k+1}^{n-2}E^{(A)}_j,\xi_{t_k}\in E^{(A)}_k \right\}.
\end{eqnarray*}
The mean first hitting time (to the set $\bigcup_{j=k+1}^{n-2}E^{(A)}_j$) of the EA starting with the population $\xi_{t_k}=X$, denoted by $\tau^{(A)}_{k,k+1}$, is given by
\begin{eqnarray*}
\mathbb{E}\left[\tau^{(A)}_{k,k+1}\mid \xi_{t_k}=X\right]=\sum_{t=0}^\infty t \mathbb{P}\left[\tau^{(A)}_{k,k+1}=t\mid \xi_{t_k}=X\right].
\end{eqnarray*}
Let $\mu_{t_k}(\cdot)$ specify the probability distribution of the population at the $t_k^{th}$ generation (note that we have assumed that $\xi_{t_k}\in E^{(A)}_k$), then
\begin{eqnarray*}
\mathbb{E}\left[\tau^{(A)}_{k,k+1}\right]=\sum_{Y\in E^{(A)}_k}\mu_{t_k}(Y) \mathbb{E}\left[\tau^{(A)}_{k,k+1}\mid \xi_{t_k}=Y\right]
\end{eqnarray*}
is called the mean first hitting time to population set $\bigcup_{j=k+1}^{n-2}E^{(A)}_j$. Now we utilize drift analysis to bound $\mathbb{E}\left[\tau^{(A)}_{k,k+1}\right]$. First, we define a distance function $V^{(k)}(X)$ for $X\in E^{(A)}_k\cup \bigcup_{j=k+1}^{n-2}E^{(A)}_j$ ($\rho_1\le k\le \rho_2\le\ln^2n$):
\begin{eqnarray*}
V^{(k)}(X) =\left\{
\begin{array}{llll}
   0, &\forall X \in \bigcup_{j=k+1}^{n-2}E^{(A)}_j; \\
   1/u_{k,\epsilon}^{(A)}+\bar{\eta}^{(A)}_{k,\epsilon}(X), &\forall X \in E^{(A)}_k.
  \end{array}
\right.
\end{eqnarray*}
For each population set $E^{(A)}_k$ ($k=1,\dots,n$), we show that the one step mean drift of the populations is no less than some positive constant. Given a population $X\in E^{(A)}_k$, let $\Delta(X)$ be its one step mean drift. The estimation of one step mean drift involves two different cases (Eq. \ref{deta1} and Eq. \ref{deta2}).

For $X \in \left(\bigcup_{i=\lceil\epsilon N\rceil}^{N} E^{(A)}_{k,i}\right),Y\in E^{(A)}_k\cup \left(\bigcup_{j=k+1}^{n-2}E^{(A)}_j\right)$, {\small
\begin{eqnarray}
  \nonumber\Delta(X) &=& \sum_{Y \in \bigcup_{j=k+1}^{n-2}E^{(A)}_j} \left[V^{(k)}(X)-V^{(k)}(Y)\right]
  \mathbb{P}(X,Y)+\sum_{Y\in E^{(A)}_k} \left[V^{(k)}(X)-V^{(k)}(Y)\right] \mathbb{P}(X,Y)
  \\\label{deta1}
&\ge & 1 -\sum_{Y \in \bigcup_{i=1}^{\lceil\epsilon N\rceil-1} E^{(A)}_{k,i}} \bar{\eta}^{(A)}_{k,\epsilon}(Y) \mathbb{P}(X,Y)\\
&\ge & 1 -\max_{Y \in \bigcup_{i=1}^{\lceil\epsilon N\rceil-1} E^{(A)}_{k,i}}\left\{\mathbb{P}(X,Y)\right\} \cdot\sum_{Y \in \bigcup_{i=1}^{\lceil\epsilon N\rceil-1} E^{(A)}_{k,i}} \bar{\eta}^{(A)}_{k,\epsilon}(Y) =1
\end{eqnarray}}
holds since the truncation selection operator always preserves the best $N$ individuals among the $2N$ individuals in the parent and offspring populations of each generation.

For $X \in \left(\bigcup_{i=1}^{\lceil\epsilon N\rceil-1} E^{(A)}_{k,i}\right) , Y\in E^{(A)}_k\cup \left(\bigcup_{j=k+1}^{n-2}E^{(A)}_j\right)$, {\small
\begin{eqnarray}\
  \nonumber\Delta(X) &=& \sum_{Y \in \bigcup_{j=k+1}^{n-2}E^{(A)}_j} \left[V^{(k)}(X)-V^{(k)}(Y)\right]
  \mathbb{P}(X,Y)+\sum_{Y \in E^{(A)}_k} \left[V^{(k)}(X)-V^{(k)}(Y)\right] \mathbb{P}(X,Y)  \\
   \nonumber&\ge & \left(\sum_{Y \in \bigcup_{j=k+1}^{n-2}E^{(A)}_j}  +\sum_{Y \in \bigcup_{i=\lceil\epsilon N\rceil}^{N}
E^{(A)}_{k,i}}\right) \bar{\eta}^{(A)}_{k,\epsilon}(X)
 \mathbb{P}(X,Y)+ \sum_{Y \in \bigcup_{i=1}^{\lceil\epsilon N\rceil-1}
E^{(A)}_{k,i}} \left[\bar{\eta}^{(A)}_{k,\epsilon}(X)-
\bar{\eta}^{(A)}_{k,\epsilon}(Y)\right] \mathbb{P}(X,Y) \\
  \nonumber&= & \sum_{Y \in \bigcup_{i=\lceil\epsilon N\rceil}^{N}
E^{(A)}_{k,i}} \bar{\eta}^{(A)}_{k,\epsilon}(X) \bar{\mathbb{P}}(X,Y)+ \sum_{Y \in \bigcup_{i=1}^{\lceil\epsilon N\rceil-1} E^{(A)}_{k,i}} \left[\bar{\eta}^{(A)}_{k,\epsilon}(X)-
\bar{\eta}^{(A)}_{k,\epsilon}(Y)\right] \mathbb{P}(X,Y) \\
 \label{deta2}       &=& \sum_{Y \in E^{(A)}_k}
        \left[\bar{\eta}^{(A)}_{k,\epsilon}(X)-\bar{\eta}^{(A)}_{k,\epsilon}(Y)\right]
        \bar{\mathbb{P}}(X,Y)=1,
\end{eqnarray}}
where Eq. \ref{deta2} holds because of Eq. \ref{m}. The results presented in Eq. \ref{deta1} and Eq. \ref{deta2} show that the one step mean drift of $X$ ($X\in E^{(A)}_k$) is always no less than 1. According to Lemma \ref{driftcondition} and Eq. \ref{eq:max_takeover}, we further have
\begin{eqnarray*}
\mathbb{E}\left[\tau^{(A)}_{k,k+1}\mid \xi_{t_k}=X\right]=1/u_{k,\epsilon}^{(A)}+\bar{\eta}^{(A)}_{k,\epsilon}(X)= O\Big(1/u_{k,\epsilon}^{(A)}+\bar{\eta}^{(A)}_{\epsilon}\Big).
\end{eqnarray*} Then the
first hitting time from $E^{(A)}_k$ to $E^-_{k}$ satisfies:
\begin{eqnarray*}
\mathbb{E}\left[\tau^{(A)}_{k,k+1}\right]&=&\sum_{X\in E^{(A)}_k}\mu_{t_k}(X) \mathbb{E}\left[\tau^{(A)}_{k,k+1}\mid \xi_{t_k}=X\right]=O\Big(1/u_{k,\epsilon}^{(A)}+\bar{\eta}^{(A)}_{\epsilon}\Big)\sum_{X\in
E^{(A)}_k}\mu_{t_k}(X)\\
&=& O\Big(1/u_{k,\epsilon}^{(A)}+\bar{\eta}^{(A)}_{\epsilon}\Big).
\end{eqnarray*}
For $X \in E^{(A)}_{k'}$ ($k'\neq k, \rho_1\le k'\le \rho_2\le\ln^2n$), we can obtain the same upper bound by the same techniques. In the meantime, the truncation selection operator of the EA always preserves the best individual in every generation (ESS), thus once the population of the EA has reached $\bigcup_{j=k+1}^{n-2}E^{(A)}_j$, it will never return to $E^{(A)}_k$ again. Hence, combining the conditional means of $\tau^{(A)}_{k,k+1}$ with respect to different $k$, we have proven the first part of Lemma \ref{lemma:unimodel_A}:
\begin{eqnarray*}
\mathbb{E}\left[\tau^{(A)}_{\rho_1,\rho_2}\bigg|\bigwedge_{t=1}^{\tau^{(A)}_{\rho_1,\rho_2}}(\mathcal{S}_0\cap\xi_{t+1}=\emptyset)\right]= O\left[\sum_{k=\rho_1+1}^{\rho_2}\left(\bar{\eta}^{(A)}_{\epsilon}+1/u_{k,\epsilon}^{(A)}\right)\right];
\end{eqnarray*}

The second part of Lemma \ref{lemma:unimodel_A} (Eq. \ref{Eq:lemma:A1}) can be proven in a similar way, except that the condition ``no individual belonging to $\mathcal{S}_0$ has ever been generated before the first time the EA reaches $E_{\rho_2}^{(A)}$'' is no longer required. $\square$
\subsection*{Proof of Lemma \ref{lemma:unimodel_B}}
\textbf{\emph{Proof.}} The idea of the proof is straightforward: in the worst case, the $(N+N)$ EA should spend $\rho_2-\rho_1$ repeated takeover-upgrade processes in which no individual belonging to $\mathcal{S}^*$ has ever been generated before the first time the EA reaches $E_{\rho_2}^{(B)}$. Instead of considering the mean of the first hitting time from $E_{\rho_1}^{(B)}$ to $E_{\rho_2}^{(B)}$, we consider the upper bound of the first hitting time which holds with an overwhelming probability, i.e., a probability that is super-polynomially close to $1$. Since the truncation selection is adopted by the EA, such an upper bound can be obtained by combining directly the upper bounds of the takeover and upgrade times together. As it is defined, with an overwhelming probability, $\hat{\eta}^{(B)}_{\epsilon}$ is an upper bound of the $(B,i,\epsilon)$-takeover time $\eta^{(B)}_{i,\epsilon}(\xi_{t_i})$ for any population $\xi_{t_i}\in E_i^{(B)}$, where $i\in\{0,\dots,n-2\}$ holds.

In addition, to prove the lemma, we have to prove that $\ln^3n$ is an upper bound of the upgrade time with an overwhelming probability. Linking the above proposition to a variable $X$ obeying the geometric distribution with parameter $u_{i,\epsilon}^{(B)}$, the probability that $X$ is bounded from above by $\ln^3n$ can be calculated as follows:
\begin{eqnarray}
\nonumber \mathbb{P}\left[X\le \ln^3n\right]&=&\sum_{k=1}^{\ln^3n}
\mathbb{P}[X=k]=1-\sum_{\ln^3n+1}^{+\infty}\mathbb{P}[X=k]\\
\label{eq:upgrade_p_lower_bound}&=&1-\sum_{k=\ln^3n+1}^{+\infty}\left(1- u_{i,\epsilon}^{(B)}\right)^{k-1}u_{i,\epsilon}^{(B)} =1-\left(1-u_{i,\epsilon}^{(B)}\right)^{\ln^3n}.
\end{eqnarray}
Concerning the upgrade probability $u_{i,\epsilon}^{(B)}$, we can estimate its lower bound under the condition $N=\Omega(n/\ln n)$. Given the population $\xi_{t_\rho}\in \left(\cup_{i=\epsilon N}^{N} E^{(A)}_{\rho,i}\right)$ containing at least $\epsilon N$ LOIBs, $u_{i,\epsilon}^{(B)}$ is bounded from below:
\begin{eqnarray*}
u_{i,\epsilon}^{(B)}\ge 1-\left[1-\frac{1}{n}\left(1-\frac{1}{n}\right)^{n-1}\right]^{\epsilon N}\ge 1-\left(1-\frac{1}{e^2n}\right)^{\epsilon N}\ge1-e^{-\frac{\epsilon N}{e^2n}}.
\end{eqnarray*}
Noting that the population size $N$ satisfies $N=\Omega(n/\ln n)$, the above inequality implies
\begin{eqnarray*}
u_{i,\epsilon}^{(B)}\ge1-e^{-\frac{\epsilon N}{e^2n}}=1-e^{-\Omega(1/\ln n)}=\Omega\left(\frac{1}{\ln n}\right).
\end{eqnarray*}
By inserting the above inequality into Eq. \ref{eq:upgrade_p_lower_bound}, we obtain
\begin{eqnarray}
\label{eq:geo_dis_prob}\mathbb{P}\left[X\le \ln^3n\right]=1-\left(1-u_{i,\epsilon}^{(B)}\right)^{\ln^3n}=1-\left[1-\Omega\left(\frac{1}{\ln n}\right)\right]^{\ln^3n}\ge 1-e^{-\Omega(\ln^2n)},
\end{eqnarray}
which is super-polynomially close to $1$. Given the population $\xi_{t_i}\in \left(\cup_{k=\epsilon N}^{N} E^{(A)}_{i,k}\right)$ at the $t_i^{th}$ generation ($\rho_1\le i\le \rho_2$), the $(B,i,\epsilon)$-upgrade time be $\phi^{(B)}_{i,\epsilon}(\xi_{t_\rho})$ obeys the geometric distribution with parameter $u_{i,\epsilon}^{(B)}$. Hence, by replacing the variable $X$ with $\phi^{(B)}_{i,\epsilon}(\xi_{t_i})$ in Eq. \ref{eq:geo_dis_prob}, we know that $\phi^{(B)}_{i,\epsilon}(\xi_{t_\rho})$ is bounded from above by $\ln^3n$ with an overwhelming probability. Noting that $\forall i\in\{0,\dots,n-2\}, \forall\xi_{t_i}\in E_i^{(B)}: \mathbb{P}\left( \eta^{(B)}_{i,\epsilon}(\xi_{t_i})\le \hat{\eta}^{(B)}_{\epsilon} \right)\succ 1-1/SuperPoly(n)$ (condition of Lemma \ref{lemma:unimodel_B}), following the proof idea presented at the beginning of the proof, we have proven that the upper bound of the first hitting time, shown in Eq. \ref{eq:upper_bound}, holds with an overwhelming probability, i.e., a probability that is super-polynomially close to $1$. $\hfill\square$
\subsection*{Proof of Lemma \ref{lemma:N+N_cheby}}
 Let $Z_{t+1}$ be the number of generated LOIAs at the $(t+1)^{th}$ generation. Given that $c\in(0,1)$ is a constant,
 for $X_t<N/(1+ch)$ we have
\begin{eqnarray}
\nonumber&&\mathbb{P}\bigg[Z_{t+1}>c\cdot \mathbb{E} [Z_{t+1}|X_t, \mathcal{S}_0\cap\xi_{t+1}=\emptyset]\Big|
X_t<\frac{N}{1+ch}\bigg]\\
\nonumber&=&\mathbb{P}\bigg[\mathbb{E} [Z_{t+1}|X_t, \mathcal{S}_0\cap\xi_{t+1}=\emptyset]-Z_{t+1}<(1-c)\cdot \mathbb{E} [Z_{t+1}|X_t, \mathcal{S}_0\cap\xi_{t+1}=\emptyset]\Big|
X_t<\frac{N}{1+ch}\bigg]\\
\nonumber&>& \mathbb{P}\bigg[\Big|\mathbb{E} [Z_{t+1}|X_t, \mathcal{S}_0\cap\xi_{t+1}=\emptyset]-Z_{t+1}\Big|<(1-c)\cdot \mathbb{E} [Z_{t+1}|X_t, \mathcal{S}_0\cap\xi_{t+1}=\emptyset]\bigg|
X_t<\frac{N}{1+ch}\bigg]\\
\nonumber&=& 1-\mathbb{P}\bigg[\Big|Z_{t+1}-\mathbb{E} [Z_{t+1}|X_t, \mathcal{S}_0\cap\xi_{t+1}=\emptyset]\Big|\ge(1-c)\cdot \mathbb{E} [Z_{t+1}|X_t, \mathcal{S}_0\cap\xi_{t+1}=\emptyset]\bigg|
X_t<\frac{N}{1+ch}\bigg]\\
\label{equation:need_to_explain}&\ge& 1-\frac{\text{Var}[Z_{t+1}|X_t<\frac{N}{1+ch}, \mathcal{S}_0\cap\xi_{t+1}=\emptyset]}{(1-c)^2\mathbb{E}^2[Z_{t+1}|X_t<\frac{N}{1+ch}, \mathcal{S}_0\cap\xi_{t+1}=\emptyset]} =1-\frac{X_th(1-h)}{(1-c)^2X_t^2h^2}=1-\frac{1-h}{(1-c)^2X_th},
\end{eqnarray}
where in Eq. \ref{equation:need_to_explain} we utilize the fact that the newly generated LOIAs can all be accepted\footnote{In our worst-case analysis, we ignore the generation of advanced LOIAs if the number of LOIAs has not reached a predefined level \cite{chen09unimodal}} (where $X_t<N/(1+ch)$ holds), the number of newly generated LOIAs obeys the binomial distribution \cite{Papoulis91book}. Hence, by considering the total number of LOIAs at the end of the $(t+1)^{th}$ generation, we know that
\begin{eqnarray}
\mathbb{P}\bigg[X_{t+1}> (1+ch)X_t\Big| X_t<\frac{N}{1+ch}, \mathcal{S}_0\cap\xi_{t+1}=\emptyset \bigg]> 1-\frac{1-h}{(1-c)^2X_th}
\end{eqnarray}
holds, where $c\in(0,1)$ is a constant.$\hfill\square$
\section*{Acknowledgement}
Tianshi Chen would like thank Dr. Jun He for his kind help over the years, and he is also grateful to Dr. Per Kristian Lehre for discussing the analysis of takeover time. This work is partially supported by two Natural Science Foundation of China grants (No. 61003064 and No. U0835002), National S\&T Major Project (Grant 2010ZX01036-001-002), the Fund for Foreign Scholars in University Research and Teaching Programs (Grant No. B07033), and an Engineering and Physical Science Research Council grant in UK (No. EP/D052785/1).

\end{document}